\documentclass[conference]{IEEEtran}
\IEEEoverridecommandlockouts
\usepackage{cite}
\usepackage{amsmath,amssymb,amsfonts}
\usepackage{algorithmic}
\usepackage{graphicx}
\usepackage{booktabs}
\usepackage{ulem}
\usepackage{multirow}
\usepackage[hyphens]{url}
\usepackage{textcomp}
\usepackage{booktabs}
\usepackage{comment}
\usepackage{xcolor}
\def\BibTeX{{\rm B\kern-.05em{\sc i\kern-.025em b}\kern-.08em
    T\kern-.1667em\lower.7ex\hbox{E}\kern-.125emX}}

\begin{document}

\DeclareRobustCommand*{\IEEEauthorrefmark}[1]{%
    \raisebox{0pt}[0pt][0pt]{\textsuperscript{\footnotesize\ensuremath{#1}}}}

\title{RNTrajRec: Road Network Enhanced Trajectory Recovery with Spatial-Temporal Transformer (Extended Version)}

\author{\IEEEauthorblockN{Yuqi Chen\IEEEauthorrefmark{1}, Hanyuan Zhang\IEEEauthorrefmark{2}, Weiwei Sun\IEEEauthorrefmark{2}, Baihua Zheng\IEEEauthorrefmark{3}}
\IEEEauthorblockA{\IEEEauthorrefmark{1,2} School of Computer Science \& Shanghai Key Laboratory of Data Science, Fudan University \\
%\IEEEauthorrefmark{1,2} Shanghai Key Laboratory of Data Science, Fudan University \\
\IEEEauthorrefmark{1,2} Shanghai Institute of Intelligent Electronics \& Systems \\
\IEEEauthorrefmark{3} School of Computing and Information Systems, Singapore Management University  \\
\IEEEauthorrefmark{1}chenyuqi21@m.fudan.edu.cn \IEEEauthorrefmark{2}\{zhanghy20, wwsun\}@fudan.edu.cn \IEEEauthorrefmark{3}bhzheng@smu.edu.sg}}

\maketitle

\begin{abstract}

%Trajectory recovery is a fundamental task 
GPS trajectories are the essential foundations for many trajectory-based applications, such as travel time estimation, traffic prediction and trajectory similarity measurement. Most applications require a large number of high sample rate trajectories to achieve a good performance. 
%The performance of these models depends largely on the high sample rate of trajectories precisely matched to the map. 
However, many real-life trajectories are collected with low sample rate due to energy concern or other constraints.
%from the GPS devices. 
We study the task of trajectory recovery in this paper as a means to increase the sample rate of low sample trajectories. 
Currently, most existing works on trajectory recovery follow a sequence-to-sequence diagram, with an encoder to encode a trajectory and a decoder to recover real GPS points in the trajectory. However, these works ignore the topology of road network and only use grid information or raw GPS points as input. Therefore, the encoder model is not able to capture rich spatial information of the GPS points along the trajectory, making the prediction less accurate and lower spatial consistent. In this paper, we propose a road network enhanced transformer-based framework, namely RNTrajRec, for trajectory recovery. RNTrajRec first uses a graph model, namely GridGNN, to learn the embedding features of each road segment. It next develops a spatial-temporal transformer model, namely GPSFormer, to learn rich spatial and temporal features along with a Sub-Graph Generation module to capture the spatial features for each GPS point in the trajectory. It finally forwards the outputs of encoder model to a multi-task decoder model to recover the missing GPS points. 
%\baihua{Shall we add one more sentence to describe the final decoding process to complete the trajectory recovery?}
Extensive experiments based on three large-scale real-life trajectory datasets confirm the effectiveness of our approach. 

\end{abstract}


\section{Introduction}
\label{sec:intro}

GPS trajectories are the essential foundations of
%have emerged for 
many applications such as travel time estimation~\cite{zhang2018deeptravel, li2018multi}, traffic prediction~\cite{li2021traffic, zheng2020gman, jin2022gridtuner}, trajectory similarity measurement~\cite{li2018deep,yao2019computing,yang2021t3s, zhang2020trajectory, han2021graph} and etc. 
In order to achieve good performances, most of these applications require a large number of high sample rate trajectories~\cite{ren2021mtrajrec}, as trajectories of low sample rate lose detailed driving information and increase uncertainty.
%
%However, the performance of these models depend highly on the high sample rate of the trajectories. Since low-sample trajectories lose detailed driving information and increase uncertainty, these models may tend to be less effective\cite{ren2021mtrajrec}. 
%
However, as pointed out in previous works~\cite{ren2021mtrajrec,zhao2019deepmm}, a large number of  trajectories generated in real-life have low sample rate, e.g., taxis usually report their GPS locations every $2 \sim 6$ minutes to reduce energy consumption~\cite{yuan2010interactive}. Consequently, it is hard for most existing models developed for the applications mentioned above to utilize these trajectories effectively. In addition, GPS trajectories have to be first mapped to the road network via map matching before being used by many applications. Most existing map matching algorithms are based on Hidden Markov Model (HMM)~\cite{newson2009hidden} and its variants, and they can achieve a high accuracy only when the trajectories are sampled in a relatively high rate~\cite{lou2009map}. Although some existing works aim to increase the accuracy of map matching, including HMM-based methods~\cite{jagadeesh2017online, song2012quick} and learning based methods~\cite{lou2009map, zhao2019deepmm}, we have not yet found a good solution to address the issues caused by low-sample trajectories.

Trajectory recovery, that aims to increase the sample rate via recovering the missing points of a given trajectory, enriches low-sample trajectories from a different perspective. We can assume that vehicles are moving with the uniform speeds and insert new points (generated by linear interpolation) between every two consecutive GPS points in the input trajectory~\cite{hoteit2014estimating}. Though easy for implementation, it suffers from poor accuracy. Recently, two learning-based methods have been developed for trajectory recovery, including DHTR~\cite{wang2019deep} and MTrajRec~\cite{ren2021mtrajrec}. 
%Typically, most existing work on trajectory recovery
Both methods follow sequence-to-sequence~\cite{sutskever2014sequence} diagram, with an encoder model to generate the representation of a given trajectory and a decoder model to recover the trajectory point by point.
%% - skip to save space
%Specifically, DHTR focuses on the recovery of the missing GPS points in a trajectory via a learning-based method and proposes a kalman filter (KF)~\cite{kalman1960new} component to reduce the uncertainty and noise of the trajectory. It then relies on existing map matching algorithms to generate the recovered trajectory. Different from DHTR, MTrajRec~\cite{ren2021mtrajrec} develops a novel end-to-end model. It proposes a map constraint layer to provide fine-grained information for each GPS point in the trajectory and a multi-task sequence-to-sequence model that is able to recover the trajectory more efficiently. As compared with DHTR, MTrajRec achieves a better performance. 
%% -end of skip
%Other related works include AttnMove~\cite{xia2021attnmove} and Bi-STDDP~\cite{xi2019modelling} that use multiple attention mechanisms and history trajectories to recover missing check-in data respectively, but they cannot be directly applied to GPS trajectories. 
Nevertheless, all existing works still suffer from following two major limitations. 
%
%The main contribution of MTrajRec is the proposed map constraint layer which aims to provide fine-grained information for each GPS point in the trajectory and a multi-task sequence-to-sequence model that solve the problem of trajectory recovery efficiently. While MTrajRec achieves a high performance as compared to two-stage solution (i.e. DHTR), there are two major problems for these existing models: 
\begin{itemize}
    \item Most of these existing works ignore the road network structure, making the prediction lack spatial consistency to a certain degree. 
    \item Most of these existing works use a simple encoder model to represent trajectories and hence are unable to fully utilize the rich contextual information of GPS trajectories. For example, MTrajRec only employs a simple gated recurrent unit (GRU)~\cite{cho2014properties} for trajectory representation.
\end{itemize}

\begin{figure}[htbp]
\centering
\includegraphics[width=7.4cm]{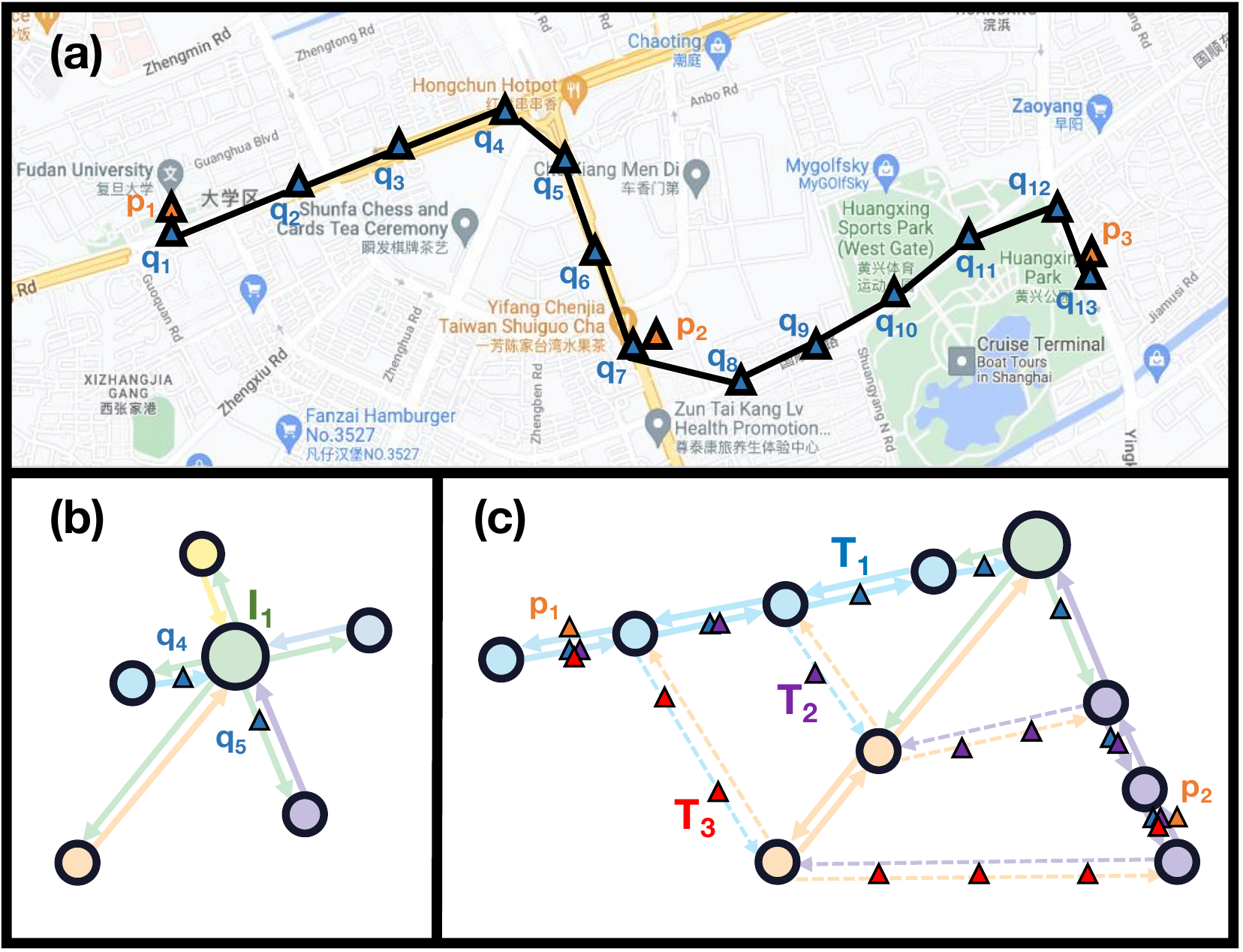}%{fig/Fig1-v2.png}%{fig/total.png}
\vspace{-0.08in}
\caption{An example of trajectory recovery task and detailed illustration to demonstrate the significance of road network structure.}
\label{fig1}
\vspace{-0.25in}
\end{figure}

To facilitate the understanding of the above two limitations, we plot an example in Fig.~\ref{fig1}(a). The three orange-colored points labelled as $p_i$ ($i\in [1,3]$) form an input trajectory of low sample rate. As observed, the distance between each two consecutive points is relatively long (as there are many missing GPS points between them). The task of trajectory recovery is to recover the missing points and map all the GPS points (including the input points and recovered points) to the road network to generate a high-sample trajectory, e.g., the 13 blue-colored points labelled as $q_j$ ($j\in [1,13]$) together with the underlying road segments (represented by black color lines) shown in Fig.~\ref{fig1}(a) form an output trajectory. 
%Fig. ~\ref{fig1.1} shows a example of trajectory recovery. Given a low-sample trajectory (marked orange in the figure), it aims to recover the missing GPS points and map each GPS point onto the road network to obtain the real GPS location of the moving trajectory (marked blue in the figure). 

%Though the task looks simple for a human being with good knowledge of the underlying road network, it 
This task is not trivial. 
%When the sample rate is low, two consecutive GPS points in the input trajectory may be located at two different road segments while there are multiple candidate routes available to connect them. 
Both the road network structure and context information of the input trajectory are essential to guarantee the high accuracy of the recovered trajectories. Their significance lies in four aspects. 
i) When two newly generated consecutive points are located at two different road segments, we have to rely on the road network to recover the trajectory. As illustrated in Fig.~\ref{fig1}(b), new points $q_4$ and $q_5$ are near a major intersection $I_1$ that has complex topology. It's impossible to recover the underlying road segments if the road network structure is not considered. Note in Fig.~\ref{fig1}(b) and (c), circles represent intersections and lines with arrows stand for directional road segments from one intersection to another.
ii) Raw GPS points by nature have errors (e.g., most GPS-enabled smartphones are accurate to within a 4.9 m radius under open sky). As shown in Fig.~\ref{fig1}(a), raw GPS points $p_1$, $p_2$ and $p_3$ are not located at any road segments. Consequently, underlying road network provides an effective means to correct the errors as vehicles must move along the road network. 
iii) Raw points in the input trajectory might be far away from each other, and underlying road network reveals how vehicles can move from a point to another, e.g., we cannot simply use a straight line to connect $p_1$ to $p_2$ in Fig.~\ref{fig1}(a).
iv) Two consecutive GPS points in the input trajectory may be located at two different road segments while there are multiple candidate routes available to connect them. For example, we plot three candidate routes from $p_1$ to $p_2$ in Fig.~\ref{fig1}(c), where triangles of the same color form one candidate route. The contextual information of the trajectory provides useful clues when filtering out impossible candidate route(s). For example, the position of $p_3$ 
%(the point sampled right after $p_2$ in the input trajectory)
suggests that the candidate route $T_3$ is unlikely to be the one as it requires detour to travel from $p_2$ to $p_3$. Those familiar with map matching might notice that some aspects discussed above are applicable to map matching too. Map matching could be considered as a sub-step of trajectory recovery, while trajectory recovery is more challenging as it not only maps GPS points to road segments but also recovers missing GPS points that truly capture a vehicle's movement. 

To address the limitations of existing trajectory recovery methods and meanwhile take advantage of end-to-end framework, 
we propose a novel transformer-based model, 
%for GPS trajectory representation, 
namely RNTrajRec, a \textbf{R}oad \textbf{N}etwork enhanced \textbf{Traj}ectory \textbf{Rec}overy framework with spatial-temporal transformer. To capture the road network structure, RNTrajRec first develops a grid-partitioned road network representation module, namely GridGNN, to learn the hidden-state embedding for each road segment. To capture both the spatial-temporal features and the contextual information of trajectory, RNTrajRec then develops a novel transformer-based model, namely GPSFormer, which first represents each GPS point in a trajectory as a sub-graph road network that surrounds the GPS point through a Sub-Graph Generation module, and then introduces a novel spatial-temporal transformer model to learn rich spatial and temporal patterns of a GPS trajectory. It finally adopts a well-designed decoder proposed in \cite{ren2021mtrajrec} on top of the encoder model to recover the missing GPS points in the trajectory. Overall, our major contributions are summarized below.
\vspace{-0.02in}
\begin{itemize}
    \item We propose a novel framework, namely RNTrajRec. 
    %, which is a \textbf{R}oad \textbf{N}etwork enhanced \textbf{Traj}ectory \textbf{Rec}overy framework with spatial-temporal transformer. 
    To the best of our knowledge, RNTrajRec is the first attempt to combine road network representation with GPS trajectory representation for the task of trajectory recovery.
    \item To consider the road network structure around each GPS point in a given trajectory, we propose a novel spatial-temporal transformer network, namely GPSFormer, which consists of a transformer encoder layer for temporal modeling and a graph transformer model, namely graph refinement layer, for spatial modeling. Meanwhile, a Sub-Graph Generation module is developed to capture the spatial features for each GPS point.
    \item We propose a novel model for road network representation, namely GridGNN, which seamlessly combines grid-level representation with road network representation.
    \item We conduct extensive experiments on three real-life datasets to compare RNTrajRec with existing methods\footnote{Codes are available at \url{https://github.com/chenyuqi990215/RNTrajRec}.}. Experimental results demonstrate that RNTrajRec significantly outperforms  state-of-the-art solutions. 
    
\end{itemize}

\section{Related Work}

%\subsection{Spatial-Temporal Transformer}
\noindent
\textbf{Spatial-Temporal Transformer.}
Transformers have achieved great success in many artificial intelligence fields, such as natural language processing~\cite{vaswani2017attention, devlin2018bert, liu2021gpt} and computer vision~\cite{carion2020end, liu2021swin, fang2021you}. Therefore, it has naturally attracted a lot of interest from academia and industry with many variants~\cite{beltagy2020longformer, zhou2021informer, kitaev2020reformer} being proposed. As a result, there is a growing interest in applying transformer architecture for graph representation\cite{yun2019graph, cai2020graph, dwivedi2020generalization}. 
%For example, Yun et al~\cite{yun2019graph} utilizes graph transformer to generate meaningful graph structures for heterogeneous graph representation; Cai et al~\cite{cai2020graph} explores relation encoder and graph transformer structure for graph-to-sequence learning; Dwivedi et al~\cite{dwivedi2020generalization} generalizes transformer architectures for arbitrary homogeneous graphs structure learning.
However, most of these works solve graph modeling on fixed graph structure, which is not suitable for modeling dynamic graphs with temporal dependency. 
%While in this paper, we regard a GPS point as a weighted sub-graph of the road network surrounding the GPS point. Therefore, the graph structure in a given sequence could be different from each other, making these work not suitable for GPS trajectory representation. 

Recently, several works have been proposed for dynamic graph modeling with transformer framework to solve pedestrian trajectory prediction\cite{yu2020spatio}, dynamic scene graph generation\cite{cong2021spatial}, 3D human pose estimation\cite{zheng20213d}, activity recognition\cite{li2021groupformer, zhang2021stst} and etc. Specifically,
Yu et al~\cite{yu2020spatio} propose a novel spatio-temporal transformer framework with intra-graph crowd interaction for trajectory prediction. However, the number of pedestrians (i.e. the number of nodes at each timestamp) is quite small and fixed. Cong et al~\cite{cong2021spatial} use the same transformer structure for both spatial and temporal modeling with a masked multi-head self-attention layer for temporal modeling. However, the time cost for each layer is $O\left(l^2 \cdot v^2 \right)$, with $l$ the length of the sequence and $v$ the averaged number of nodes in each graph structure.
%, which is not efficient enough. 
Zheng et al~\cite{zheng20213d} use a patch embedding for spatial position embedding and propose a spatial attention network. However, the input and the output have different structures and hence the model is not stackable. Li et al~\cite{li2021groupformer} propose a spatial-temporal transformer framework tailored for group activity recognition. Zhang et al~\cite{zhang2021stst} propose a spatial-temporal transformer with a directional temporal transformer block to capture the movement patterns of human posture.

In this work, we regard a GPS point as a weighted sub-graph of the road network surrounding the GPS point. Therefore, the graph structures of points along a trajectory
%in a given sequence 
could be different from each other. We propose a spatial-temporal transformer network that has the following three advantages. i) Our model is more flexible as it has zero restriction on the number of nodes or edges for each graph. ii) The time cost for each layer is $O\left(l^2+l \cdot v\right)$, which is more efficient and scalable. Note that due to the sparseness of road network, a sub-graph of a road network with $v$ road segments only has $O\left(v\right)$ edges connecting these road segments. iii) The input and the output share the same structure, therefore, our model is stackable. 

%Following that work, we use the standard transformer layer to model temporal dependency and propose a new graph refinement layer to model spatial dependency. However, the number of pedestrians (i.e. the number of nodes at each timestamp) is quite small and fixed. For GPS trajectory, the sub-graph structure for each timestamp varies greatly, i.e. the nodes and edges for each sub-graphs are different from each other, making the task of GPS trajectory representation on road network very challenging.

%\subsection{Road Network Representation Learning}
\noindent
\textbf{Road Network Representation Learning.}
Most of the existing works regard road network as a directed graph. Node2vec~\cite{grover2016node2vec} and DeepWalk~\cite{perozzi2014deepwalk} are two novel models to model road segments as shallow embeddings. With the rapid development of graph neural network (GNN), many graph convolutional networks (e.g., GCN~\cite{kipf2016semi}, GraphSage~\cite{hamilton2017inductive}, GAT~\cite{velivckovic2017graph} and GIN~\cite{xu2018powerful}) are suitable for road network representation. Recently, several works have been proposed specifically for road network representation learning~\cite{jin2021spatio, wu2020learning, chen2021robust}. Among them, STDGNN~\cite{jin2021spatio} regards road network as a dual graph structure with a node-wise GCN modeling the features of intersections and an edge-wise GCN modeling the features of road segments,  HRNR~\cite{wu2020learning} constructs a three-level neural architecture to learn rich hierarchical features of road network, and Toast~\cite{chen2021robust} proposes a traffic context aware skip-gram module for road network representation and a trajectory-enhanced transformer module for route representation.

In this work, we regard each road segment as a sequence of grids that the road segment passes through and use recurrent neural network to model grid sequence dependency and graph neural network to model graph structure. With this design, our road network representation can capture both nearby neighborhood features and graph topology features.

%\subsection{GPS Trajectory Representation Learning}
\noindent
\textbf{GPS Trajectory Representation Learning.}
Learning-based methods for representing GPS trajectory have been studied wildly these years. Among them, T2vec~\cite{li2018deep}, NeuTraj~\cite{yao2019computing}, T3S~\cite{yang2021t3s}, and Traj2SimVec~\cite{zhang2020trajectory} are representative models. T2vec proposes the first deep learning model for trajectory similarity learning with BiLSTM~\cite{hochreiter1997long} modeling temporal dependency. NeuTraj integrates a spatial-memory network for trajectory encoding and uses distance-weighted rank loss for accurate and effective trajectory representation learning. T3S uses a self-attention based network for structural information representation and a LSTM~\cite{hochreiter1997long} module for spatial information representation. Traj2SimVec proposes a novel sub-trajectory distance loss and a trajectory point matching loss for robust trajectory similarity computation. However, these works only consider GPS trajectory in Euclidean space but ignore the important road network topology. Recently, Han et al~\cite{han2021graph} propose a graph-based method along with a novel spatial network based metric similarity computation. However, it mainly focuses on representing point-of-interests (POIs) in road networks. Since POIs are discrete GPS points on road network and GPS points in a trajectory are continuous, it cannot be directly used for GPS trajectory representation learning. To the best of our knowledge, our work is the first attempt to solve GPS trajectory representation learning in spatial networks.

%\subsection{Trajectory Recovery}
\noindent
\textbf{Trajectory Recovery.}
Recovering low sample trajectory is important for reducing uncertainty~\cite{wang2019deep, xia2021attnmove, xi2019modelling, ren2021mtrajrec}. Specifically, DHTR~\cite{wang2019deep} proposes a two-stage solution that first recovers high-sample trajectory followed by a map matching algorithm (i.e. HMM~\cite{newson2009hidden}) to recover the real GPS locations. AttnMove~\cite{xia2021attnmove} designs multiple intra- and inter- trajectory attention mechanisms to capture user-specific long-term and short-term patterns. Bi-STDDP~\cite{xi2019modelling} integrates bi-directional spatio-temporal dependence and users’
dynamic preferences to capture complex user-specific patterns. However, both AttnMove and Bi-STDDP use user-specific history trajectories and are designed for recovering missing POI check-in data, which is very different from our setting. 
%which is not available in our setting.
MTrajRec~\cite{ren2021mtrajrec} proposes an end-to-end solution with a map-constraint decoder model, which significantly outperforms two-stage methods. However, it ignores the important road network structure, leaving rooms for further improving the accuracy. 
%making the prediction less accurate. 
In this paper, we propose a novel GPSFormer module to learn rich spatial and temporal features of trajectories from the road network.

\section{Preliminary}

\begin{table*}[htbp]
\renewcommand\tabcolsep{3.3pt} % 调整表格列间的宽度
\renewcommand{\arraystretch}{1.2} % Default value: 1
\caption{The Summary of notations}
\centering
\resizebox{0.96\textwidth}{!}{%
\begin{tabular}{|c|l|}
\hline
\textbf{Notation} & \textbf{Definition} \\
\hline
$G$ & The topology structure of the road network. \\
\hline
$\hat{G}_{\tau}$ & The subgraphs structure for the given trajectory $\tau$. \\
\hline
$\hat{G}_{\tau, i}$ & The subgraph structure of the $i^{th}$ sample point in the given trajectory $\tau$. \\
\hline
$\Sigma^{grid}$ & The grid embedding table. \\
\hline
$\Sigma^{road}$ & The road segment embedding table. \\
\hline
$e_i, \bar{e}, \tilde{e}, \hat{e}$ & A certain road segment in the road network. \\
\hline
$X^{road}$ & The representation of the road network. \\
\hline
$\vec{Z}_{\tau}^{(l)}$ & The representation for the subgraphs of the given trajectory $\tau$ after $l$ layers of GPSFormer. \\
\hline
$\vec{Z}_{\tau, i}^{(l)}$ & The representation for the subgraph of the $i^{th}$ sample in the given trajectory $\tau$ after $l$ layers of GPSFormer. \\
\hline
$\vec{H}_{\tau}^{(l)}$ & The input features for the given trajectory $\tau$ to the transformer encoder layer at the $l^{th}$ layer of GPSFormer. \\
\hline
$H_{\tau}^{traj}$ & The final representation for every sample point in the given trajectory $\tau$. \\
\hline
$h_{\tau}^{traj}$ & The final trajectory-level representation for the given trajectory $\tau$. \\
\hline
$c$ & The constraint mask vector defined in the decoder model. \\
\hline
$\mathcal{L}_{id},\mathcal{L}_{rate},\mathcal{L}_{enc}$ & The loss functions employed in RNTrajRec. \\
\hline
\end{tabular}}
\label{tab5}
\end{table*}

In this section, we formally introduce the key concepts related to this work and define the task of trajectory recovery. The notations used in this paper are briefly summarized in Table \ref{tab5} and fully explained throughout the paper.

%\yuqi{Should we simplify some definitions as suggested by the Review 6?}\baihua{We can. I combine Definition 2/3/4 into one. See whether it makes sense. }\yuqi{I think it make sense.}

\textit{Definition 1:} (\textbf{Road Network}.) A \textit{Road Network} is modeled as a directed graph $G=\left(V,E\right)$, where $V$ represents the set of road segments and $E \subseteq V \times V$ captures the connectivity of these road segments, i.e., an edge $\langle e_i, e_j \rangle \in E$ if and only if there exists a direct connection from road segment $e_i$ to road segment $e_j$.
%nodes and the set of edges respectively. In this work, we regard road segments as nodes and the connectivity of these road segments as edges. Specifically, an edge $\langle e_i, e_j \rangle \in E$ if and only if there exists a direct connection from road segment $e_i$ to road segment $e_j$. 
%For simplicity, we assume road segments are straight lines and a road segment that is in curve shape in reality can be modelled as multiple shorter road segments of straight lines.

%\textit{Definition 2:} (\textbf{GPS Trajectory}.) A \textit{GPS Trajectory} $\tau$ is defined as a sequence of $l_\tau$ tuples, i.e. $\tau=\langle \left(p_1, t_1\right), \left(p_2, t_2\right), \cdots, \left(p_{l_\tau}, t_{l_\tau}\right) \rangle$, where $l_\tau$ refers to the length of the trajectory, $p_i$ in the form of $\left(lat_i, lng_i \right)$ refers to the $i^{th}$ sample point in $\tau$ with latitude $lat_i$ and longitude $lng_i$, and $t_i$ is the sample time of $p_i$ in $\tau$. We also define $p_i=\left(lat_i, lng_i\right)$, therefore, a \textit{GPS Trajectory} $\tau$ can be simplified as $\tau=\langle \left(p_1, t_1\right), \left(p_2, t_2\right) ... \left(p_n, t_n\right) \rangle$. 

\textit{Definition 2:} (\textbf{Trajectory}.) A \textit{Trajectory} $\tau/\rho$ is defined as a sequence of $l$ tuples, i.e. $\tau / \rho = \langle \left(p_1, t_1\right)$, $\left(p_2, t_2\right)$, $\cdots$, $\left(p_{l}, t_{l}\right) \rangle$, where $l$ refers to the length of the trajectory,  $p_i$ refers to the $i^{th}$ sample point in $\tau$/$\rho$, and $t_i$ is the sample time of $p_i$ in $\tau$/$\rho$. 
Note, the time interval between two adjacent sample points in a trajectory (i.e., $t_{i+1}-t_i$) defines the sample interval $\epsilon$ of this trajectory.

GPS devices have measurement errors. Throughout the paper, we use terms \textbf{Raw GPS Trajectory} and \textbf{Map-matched GPS Trajectory} to refer to a GPS trajectory (denoted as $\tau$) directly obtained from certain GPS device and that (denoted as $\rho$) after running a map matching algorithm (e.g. HMM~\cite{newson2009hidden}) respectively.
%and use term \textbf{Map-matched GPS Trajectory} (denoted as $\rho$) to represent the GPS trajectory after running a map matching algorithm (e.g. HMM~\cite{newson2009hidden}). 
Note, $p_i$ in a raw GPS trajectory $\tau$ records its exact location using latitude and longitude, while $p_j$ in a map-matched GPS trajectory $\rho$ captures its location based on the road segment $e_j$ that $p_j$ is located at and moving ratio $r_j\in [0,1)$ that  captures the moving distance of $p_j$ over the total length of $e_j$ (e.g., if $r_j=0.5$, the point $p_j$ is located at the middle point of road segment $e_j$). In addition, a raw GPS trajectory $\tau$ typically does not have a fixed sample interval, and we use the average time interval $\epsilon_\tau$ instead. A low-sample trajectory has a long sample interval.
%%Throughout the paper, we use $\tau$ to represent a \textit{Raw GPS Trajectory}, while use $\rho$ to represent a \textit{Map-matched GPS Trajectory.}

%\textit{Definition 3:} (\textbf{Map-matched GPS Trajectory}.) A \textit{Map-matched GPS Trajectory} $\rho$ with a predefined \textit{Road Network} $G=\left(V,E\right)$ is denoted as a sequence of $l_\rho$ tuples, i.e. $\rho=\langle \left(q_1, t_1\right), \left(q_2, t_2\right), \cdots, \left(q_{l_\rho}, t_{l_\rho}\right) \rangle$, where $l_\rho$ refers to the length of $\rho$, $q_j$ in the form of $\left(e_j, r_j\right)$ indicates the road segment $e_j$ that the $j^{th}$ sample point in $\rho$ is located at and moving ratio $r_j\in [0,1)$ captures the moving distance of the $j^{th}$ sample point over the total length of $e_j$ (e.g., if $r_j=0.5$, the $j^{th}$ point is located at the middle point of road segment $e_j$). 

%\textit{Definition 4:} (\textbf{Sample Interval}.) A \textit{Sample Interval} $\epsilon$ is defined as the time interval between two adjacent sample GPS points in a trajectory. Note that for \textit{Raw GPS Trajectory} $\tau$, its sample interval might not be fixed, and we use the average time interval $\epsilon_\tau$ to define its sample interval. A low-sample trajectory typically has a long sample interval.

\textit{Definition 3:} (\textbf{Map-matched $\epsilon_\rho$-Sample Interval Trajectory}.) A \textit{Map-matched $\epsilon_\rho$-Sample Interval Trajectory} $\rho$ is a \textit{Map-matched GPS Trajectory} with a fixed sample interval $\epsilon_\rho$, i.e., $\rho=\langle \left(q_1, t_1\right), \left(q_2, t_1+\epsilon_\rho\right), \cdots, \left(q_{l_{\rho}}, t_1+(l_{\rho}-1)\epsilon_\rho\right) \rangle$.
%where $\forall j \in [1,{l_{\rho}})$, $t_{j+1}=t_j+\epsilon_\rho$.
%i.e., a Map-matched GPS Trajectory $P=\langle \left(q_1, t_1\right), \left(q_2, t_2\right)... \left(q_m, t_m\right) \rangle$ is considered as a \textit{Map-matched $\epsilon$-Sample Interval Trajectory} if $\forall j \in [1,m), t_{j+1}-t_j=\epsilon$.

%\section{Problem Statements}
%
%In this section, we introduced the task of trajectory recovery.

\textit{Definition 4:} (\textbf{Trajectory Recovery}.) Given a low-sample \textit{Raw GPS Trajectory} $\tau$ with measurement errors (e.g., orange GPS points in Fig.~\ref{fig1}(a)), the task of \textit{Trajectory Recovery} aims to recover the real \textit{Map-matched $\epsilon_\rho$-Sample Interval Trajectory} $\rho$ (e.g., blue GPS points in Fig.~\ref{fig1}(a)). Specifically, for each low-sample trajectory, it infers the missing GPS points and maps each GPS point (including the GPS points in the input trajectory $\tau$) onto the road network to obtain the real GPS locations of the moving trajectory. Note that the sample interval $\epsilon_\rho$ of the recovered trajectory $\rho$ must be much smaller than 
%the averaged sample interval 
that of the given \textit{Raw GPS Trajectory} $\epsilon_\tau$.

\section{Methodology}

\begin{figure*}[htbp]
\centering
\includegraphics[width=18cm]%{fig/Fig2-v5.pdf}
{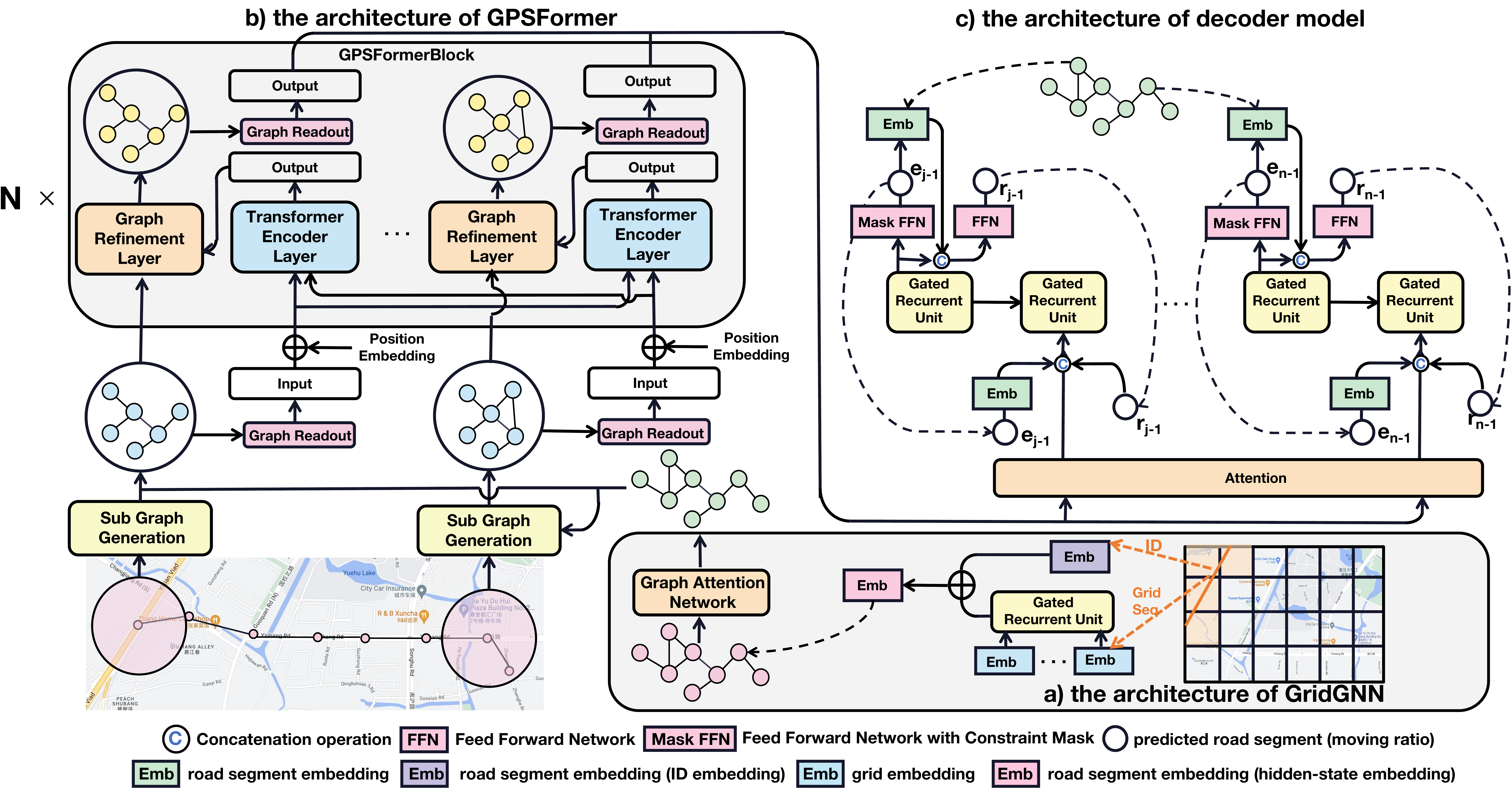}%{fig/model_new.png}
\vspace{-0.1in}
\caption{The framework of RNTrajRec. 
%which contains a grid-partition based road network representation, i.e. GridGNN, that outputs the representation of each road segment, a novel encoder model, i.e. GPSFormer, that outputs the representation of the given GPS trajectory and a well-designed decoder model from \cite{ren2021mtrajrec} that performs road segment classification and moving ratio regression for trajectory recovery task. 
(a) The architecture of GridGNN. For each road segment, a grid GRU cell is used to aggregate the grid sequence followed by a stack of $M$ GAT modules to capture the spatial information.  (b) The architecture of GPSFormer. Given a GPS trajectory, GPSFormer first extracts the road network features around each GPS point through the Sub-Graph Generation module. The features along with the sub-graphs are passed through several spatial-temporal transformer layers to learn the spatial and temporal features. (c) The architecture of decoder model. Given the output from the encoder model, an attention module is adopted to compute the similarity between the hidden-state vector of GRU cell and the outputs of encoder model and generate the input hidden-state vectors $a^{(j)}$ for timestamp $j$. A multi-task learning module is proposed that first predicts the target road segment $e_j$, followed by a regression task for predicting moving ratio on the predicted road segment, i.e. $r_j$. The entire model is trained end-to-end with Adam optimizer.}
\label{fig2}
\vspace{-0.15in}
\end{figure*}

This section introduces our proposed RNTrajRec,
%% -- skip to save space
%which consists of i) GridGNN, a grid-partitioned road network representation module that learns spatial features for each road segment; ii) GPSFormer, a spatial-temporal transformer based GPS trajectory encoder that encodes a given GPS trajectory into hidden vectors; and iii) a decoder model proposed in \cite{ren2021mtrajrec} specifically designed for trajectory recovery task, 
%%-- end of skip
with the overall framework presented in Fig.~\ref{fig2}.  

\subsection{Model Overview}

%We present our proposed framework RNTrajRec in Fig.~\ref{fig2}. 
The first component of RNTrajRec is GridGNN, a grid-partitioned road network representation module that learns spatial features for each road segment, as shown in Fig.~\ref{fig2}(a).
Given a road network $G=\left(V,E\right)$, 
%we propose a grid-partitioned road network representation module, i.e., GridGNN as shown in Fig.~\ref{fig2}(a), to 
GridGNN learns rich road network features $X^{road} \in \mathbb{R}^{|V| \times d}$, where $d$ is the hidden size of the model. 

The second component of RNTrajRec is GPSFormer, a spatial-temporal transformer based GPS trajectory encoder that encodes raw GPS points $\langle p_1, p_2, \cdots, p_
{l_\tau}\rangle$ in a trajectory $\tau$ into hidden vectors, as shown in Fig.~\ref{fig2}(b).
%%
%We design a novel representation model for encoding raw GPS points $\langle p_1, p_2, \cdots, p_{l_\tau}\rangle$ in a trajectory $\tau$, namely GPSFormer, as shown in Fig.~\ref{fig2}(b). 
%%
To obtain the input of GPSFormer, we first extract the road network features around each GPS point $p_i$ through the Sub-Graph Generation module. After the generation process, each GPS point $p_i\in \tau$ is represented as a weighted directed sub-graph $\hat{G}_{\tau, i}=\left(V_{\tau, i}, E_{\tau, i}, W_{\tau, i}\right)$, where $V_{\tau, i}$ captures the road segments selected by the module that surround the GPS point $p_i\in \tau$, $E_{\tau, i}$ is the set of edges in the selected sub-graph of the road network, and $W_{\tau, i}$ refers to the set of weights between $p_i$ and each selected road segment in the sub-graph. Note that we use $\hat{G}$ instead of $G$ to represent a sub-graph. Each generated sub-graph gathers road network features from $X^{road}$ to form its initial representation, i.e. $\vec{Z}_{\tau, i}^{(0)} \in \mathbb{R}^{|V_{\tau, i}| \times d}$. We further perform weighted mean pooling on graph to get the input representation of the trajectory $\tau$, i.e. $\hat{H}_{\tau}^{(0)} \in \mathbb{R}^{l_\tau \times d}$. 
%
%In order to get the representation of GPS trajectory, 
We then forward a mini-batch of $b$ trajectory features %$\{ H_{\tau_1}^{(0)}, H_{\tau_2}^{(0)},...,H_{\tau_b}^{(0)} \}$ and $\{ Z_{\tau_1}^{(0)},Z_{\tau_2}^{(0)},...,Z_{\tau_b}^{(0)} \}$ 
along with the sub-graph structure %$\{\hat{G}_{\tau_1}, \hat{G}_{\tau_2},..., \hat{G}_{\tau_b}\}$, with
%$\hat{G}_{\tau}=\{\hat{G}_{\tau, 1}, \hat{G}_{\tau, 2},...,\hat{G}_{\tau, n}\}$ and $Z_{\tau}^{(0)}=\{ Z_{\tau, 1}^{(0)}, Z_{\tau, 2}^{(0)},...,Z_{\tau, n}^{(0)} \}$ 
into $N$ stacked GPSFormerBlock layers, which is a combination of transformer encoder layer for temporal modeling and graph refinement layer, namely GRL, for spatial modeling. %Specifically, transformer encoder layer consists of two sub-layers, i.e. multi-head attention and feed forward, with a residual connection and layer normalization for each sub-layer, graph refinement layer also consists of two sub-layers, i.e. gated fusion and graph forward, with a residual connection and graph normalization for each sub-layer.

The last component of RNTrajRec is a decoder model, proposed in \cite{ren2021mtrajrec} specifically designed for trajectory recovery task. 
Given the outputs of a mini-batch of $b$ trajectories from the encoder model,
%\baihua{Is from encoder model or GPSFormer component/GRL?},
i.e. $\vec{H}^{(N)} \in \mathbb{R}^{b \times l_\tau \times d}$, the decoder model 
%proposed in \cite{ren2021mtrajrec} 
first uses an attention module to calculate the similarity between the hidden-state vectors of the Gated Recurrent Unit (GRU)\cite{cho2014properties} cell (i.e., the query vectors) and the outputs from the encoder model (i.e., the key vectors) to obtain the input hidden vector $a^{(j)}$ at the $j^{th}$ timestamp. Furthermore, a multi-task learning module is proposed specifically for trajectory recovery task that first predicts the target road segment $e_j$ and then predicts the corresponding moving ratio $r_j$ via a regression task, which will be detailed in Section~\ref{GCL}.  
%on the predicted road segment.

\subsection{Road Network Representation: GridGNN}

As stated in Section~\ref{sec:intro}, the road network structure is essential for the task of trajectory recovery. GridGNN is proposed to well capture the spatial features of road network. It partitions the road network into $m \times n$ equal-sized grid cells. Accordingly, each road segment can be represented as a sequence of grids that the road segment passes through.
%
%As shown in the right corner of Fig. ~\ref{fig2}, we first partition the road network into $m \times n$ grid cells. Therefore, each road segment can be represented as a sequence of grids that the road segment passes through.
%
Formally speaking, we build a grid embedding table $\Sigma^{grid} \in \mathbb{R}^{m \times n \times d}$ for each grid cell. Similarly, given the road network $G=\left(V,E\right)$, we create a road segment embedding table $\Sigma^{road} \in \mathbb{R}^{|V| \times d}$, where $|V|$ is the total number of road segments in the road network.

For each road segment $e_i\in V$, let $S_i=\langle \tilde{g}_{i}^{1}, \tilde{g}_{i}^{2},...,\tilde{g}_{i}^{\phi_i} \rangle$ be a sequence of $\phi_i$ grids passed through by $e_i$. 
%, where $\phi_i$ is the total number of grids that the road segment passes through. 
Since the grid sequence of each road segment has sequential dependencies, we use GRU cell to model the grid-level representation. That is, for each road segment $e_i$ and its corresponding grid sequence $S_i$, the grid-level hidden-state vector is given by: 
\begin{equation}
\begin{array}{rl}
    g_{i}^{(j)} = & \operatorname{lookup}(\tilde{g}_{i}^{j}.x, \tilde{g}_{i}^{j}.y) \\
    z_{i}^{(j)} = & \sigma\left(\mathbf{W}_z \cdot \left[ s_{i}^{(j-1)}, g_{i}^{(j)} \right] + \mathbf{b}_z \right) \\
    r_{i}^{(j)} = & \sigma\left(\mathbf{W}_r \cdot \left[ s_{i}^{(j-1)}, g_{i}^{(j)} \right] + \mathbf{b}_r \right) \\
    c_{i}^{(j)} = & tanh \left(\mathbf{W}_c \cdot \left[ r_{i}^{(j)} * s_{i}^{(j-1)}, g_{i}^{(j)} \right] + \mathbf{b}_c \right) \\
    s_i^{(j)} = & \left(1 - z_{i}^{(j)} \right) * s_{i}^{(j-1)} + z_{i}^{(j)} * c_{i}^{(j)} \\
\end{array}
\label{eq2}
\end{equation}
Here, $j \in \{1, 2,\cdots, \phi_{i}\}$, $\operatorname{lookup}\left(i,j\right)$ retrieves the grid embedding of position $\left(i,j\right)$ from the grid embedding table $\Sigma^{grid}$, $\mathbf{W}_x$ represents the weight for the gate($x$) neurons, $\mathbf{b}_x$ represents the bias for gate($x$), and $\sigma\left(\cdot\right)$ represents the gated function, which is implemented as sigmoid function. The initial embedding for each road segment $r_i^{(0)}$ is:
\begin{equation}
r_i^{(0)}=\operatorname{ReLU}(s_i^{(\phi_i)} + \sigma_i^{road})
\label{eq3}
\end{equation}
Here, $\sigma_i^{road} \in \mathbb{R}^{d}$ is the road segment embedding of the $i^{th}$ road segment from $\Sigma^{road}$.

The obtained hidden-state vector $r_i^{(0)}$ considers each road segment independently, which does not capture the topology of the road network structure. In order to enhance the representation of road network, we integrate GNN. Specifically, we stack $M$ layers of Graph Attention Network (GAT)
%\cite{velivckovic2017graph} 
module, which uses multi-head attention to efficiently learn the complex graph structure, to obtain final hidden-state vectors $\hat{X}^{road} \in \mathbb{R}^{|V| \times d}$:
\begin{eqnarray}
a_{ij, k}^{(l)}=&\frac{\exp \left(\operatorname{LeakyReLU}\left(\overrightarrow{\mathrm{a^k}}\left[\mathbf{\widehat{W}}^k r_{i}^{(l-1)} \| \mathbf{\widehat{W}}^k r_j^{(l-1)} \right]\right)\right)}{\sum_{n \in N_{i}} \exp \left(\operatorname{LeakyReLU}\left(\overrightarrow{\mathrm{a^k}}\left[\mathbf{\widehat{W}}^k r_{i}^{(l-1)} \| \mathbf{\widehat{W}}^k r_n^{(l-1)}\right]\right)\right)}  \label{eq5}\\
r_{i}^{(l)}=&\Vert_{k=1}^{h} \operatorname{LeakyReLU}\left(\sum\nolimits_{j \in \mathcal{N}_{i}} a_{ij,k}^{(l)} \mathbf{W}^{k} r_{j}^{(l-1)}\right) \quad \quad \ 
\label{eq6}
%\end{equation}
\end{eqnarray}
Here, $l \in \{1, 2, \cdots, M\}$. For the $k^{th}$ attention head, $a_{ij, k}^{(l)}$ represents the attention score between road segments $e_i$ and $e_j$ at the $l^{th}$ layer, $\overrightarrow{\mathrm{a}^k}$ are learnable weights to obtain attention scores, $\mathbf{W}^k$ and $\mathbf{\widehat{W}}^k$ is a learnable weight for feature transformation. $\left[\cdot\|\cdot \right]$ represents the concatenation operation, and $N_i$ represents the neighborhood of road segment $e_i$ in the road network.

The final road network representation is given by the concatenation of  $\hat{X}^{road}=\{ r_i^{(M)} \}$ for each $e_i \in V$ and the static features $f_s^{road} \in \mathbb{R}^{|V| \times f_r}$ (i.e. length of the road segment, number of in/out-going edges, etc), followed by linear transformation to obtain $d$ dimension vectors $X^{road}$.
%The final road network representation is given by $X^{road}=\{ r_i^{(M)} \}$ for each $e_i \in V$.

\subsection{Sub-Graph Generation} \label{SG}

Given the representation $X^{road} \in \mathbb{R}^{|V| \times d}$ of the road network, a straight-forward way to obtain the representation of a GPS point is to average over the embeddings of the road segments around the GPS point. However, this approach suffers from the following two problems. a) The influence of the nearby road segments on the given GPS point varies significantly. b) For each GPS point in a trajectory, its surrounding sub-graph structure is important for understanding the movement of the trajectory. Therefore, it is necessary to take the graph structure into consideration.

For a given GPS point $p$, we first 
%use R-tree\cite{guttman1984r}, an efficient data structure for spatial search, to 
locate the road segments within at most $\delta$ meters away from $p$, via R-tree\cite{guttman1984r} or any other spatial data structure. Note, $\delta$ is a hyper-parameter to control the receptive field of the GPS point. 
Assume in total $\omega_p$ road segments are returned, denoted as $\{e_{p}^1, e_{p}^2, \cdots, e_p^{\omega_p}\}$.
%
%the road segments selected by R-tree of $p$ are $V^p=\{s_{p}^1, s_{p}^2, ..., s_p^{\omega_p}\}$, where $s_{p}^i$ represents the $i$-th selected road segment and $\omega_p$ represents the number of selected road segments. 
We further follow the road network $G=(V,E)$ to connect these $\omega_p$ road segments into a sub-graph $\tilde{G}^p=\left(V^p, E^p\right)$, where $V^p=\{e_{p}^1, e_{p}^2, \cdots, e_p^{\omega_p}\}$ and $E^p = (V^p \times V^p) \cap E$. 
%and $(s_{p}^i, s_p^{j}) \in E^p$ if there exist a direct connection between road segment $s_{p}^i$ and road segment $s_p^{j}$. \baihua{can we just say: $E^p \subseteq (V^p \times V^p) \cap E$? I think yes or just $E^p = (V^p \times V^p) \cap E$}

Following \cite{ren2021mtrajrec}, we use the exponential function to model the influence of road segment $e$ on the given GPS point $p$, as defined in Eq.~\eqref{eq7}.
Here, $\operatorname{dist}(e, p)$ represents the distance between the GPS point $p$ and the road segment $e$, i.e. spherical distance between the GPS point $p$ and its projection to the road segment $e$, and $\gamma$ is a hyper-parameter with respect to the road network. To this end, we obtain the weighted sub-graph of a given GPS point, i.e. $\tilde{G}^p=\left(V^p, E^p, W^p\right)$, where $W^p$ can be derived by Eq.~\eqref{eq7}.
%
%\begin{equation}
\begin{eqnarray}
%w(\hat{e}, p) = \exp \left( \frac{-\operatorname{dist}(\hat{e}, p)^2}{\gamma^2} \right)
\omega(e, p) &=& \exp \left( {-\operatorname{dist}^2(e, p)}/{\gamma^2} \right)
\label{eq7}\\
%\end{equation}
%\begin{equation}
%g^p = \frac{\sum_{\hat{e} \in V^p} W_{\hat{e}}^p * x_{\hat{e}}^{road}}{\sum_{\hat{e} \in V^p} W_{\hat{e}}^p}
g^p &=& {\sum\nolimits_{\hat{e} \in V^p} W_{\hat{e}}^p * x_{\hat{e}}^{road}}/{\sum\nolimits_{\hat{e} \in V^p} W_{\hat{e}}^p}
\label{eq8}
%\end{equation}
\end{eqnarray}

We use mean pooling on graph to get the representation of a given GPS point $p$, as defined in Eq.~\eqref{eq8}.
Here, $x_{\hat{e}}^{road}$ represents the road segment representation of road segment $\hat{e}$ obtained from $X^{road}$ and $W_{\hat{e}}^p$ represents the influence of road segment $\hat{e}$ on the GPS point $p$, i.e. $W_{\hat{e}}^p=\omega(\hat{e}, p)$.

Given a trajectory $\tau=\langle \left(p_1, t_1\right), \left(p_2, t_2\right), \cdots, \left(p_{l_\tau}, t_{l_\tau}\right) \rangle$,
its hidden-state vector $H^{traj}_{\tau} \in \mathbb{R}^{{l_\tau} \times \left(d+3\right)}$ concatenates the road network empowered GPS representation $g_{\tau} =\langle g^{p_1}$,$g^{p_2}$, $\cdots$, $g^{p_{l_\tau}} \rangle$ $\in \mathbb{R}^{l_\tau \times d}$, timestamp sequence $\hat{t}_{\tau}=$ $\langle t_1$, $t_2$, $\cdots$, $t_{l_\tau} \rangle$ $\in \mathbb{R}^{l_\tau \times 1}$ and grid index $\hat{g}_{\tau}=$ $\langle (x_1, y_1)$, $(x_2, y_2)$, $\cdots$, $(x_{l_\tau}, y_{l_\tau}) \rangle \in \mathbb{R}^{l_\tau \times 2}$, where $(x_i, y_i)$ represents the index of the grid that the GPS point $p_i$ falls inside. Finally, we map $H^{traj}_{\tau}$ into $d$ dimension vectors $\hat{H}_{\tau}^{(0)}$ through linear transformation.
%
%\begin{equation}
%\widehat{H}_{\tau}^{(0)} = H^{traj}_{\tau} %\mathbf{W}_{l}  + \mathbf{b}_{l}
%\label{eq9}
%\end{equation}
%Here, $\mathbf{W}_{l} \in \mathbb{R}^{(d + 3) \times %d}$ and $\mathbf{b}_{l} \in \mathbb{R}^{d}$ are  learnable parameters for the linear transformation.

For the sub-graph input of the given trajectory $\tau$, $\hat{G}_{\tau}=$ $\langle \hat{G}_{\tau, 1}$, $\hat{G}_{\tau, 2}$, $\cdots, \hat{G}_{\tau, l_{\tau}} \rangle$, where $\hat{G}_{\tau, i}=\tilde{G}^{p_i}$ represents the generated weighted sub-graph for GPS point $p_i$. The initial graph-level sequence $\vec{Z}_{\tau}^{(0)} = \langle \vec{Z}_{\tau, 1}^{(0)}, \vec{Z}_{\tau, 2}^{(0)}, \cdots, \vec{Z}_{\tau, l_{\tau}}^{(0)} \rangle$, 
where $\vec{Z}_{\tau, i}^{(0)}=\{ x_{\hat{e}}^{road} \}$ for every road segment $\hat{e} \in V_{\tau, i}$.

\subsection{Graph Refinement Layer (GRL)} \label{GR}

As mentioned in Section~\ref{SG}, graph structure is significant for representing GPS points in spatial network. To this end, we propose a graph refinement layer (GRL) that is capable of capturing rich spatial features.%%, as part of GPSFormer component.

\begin{figure}[htbp]
\centerline{\includegraphics[width=8cm]{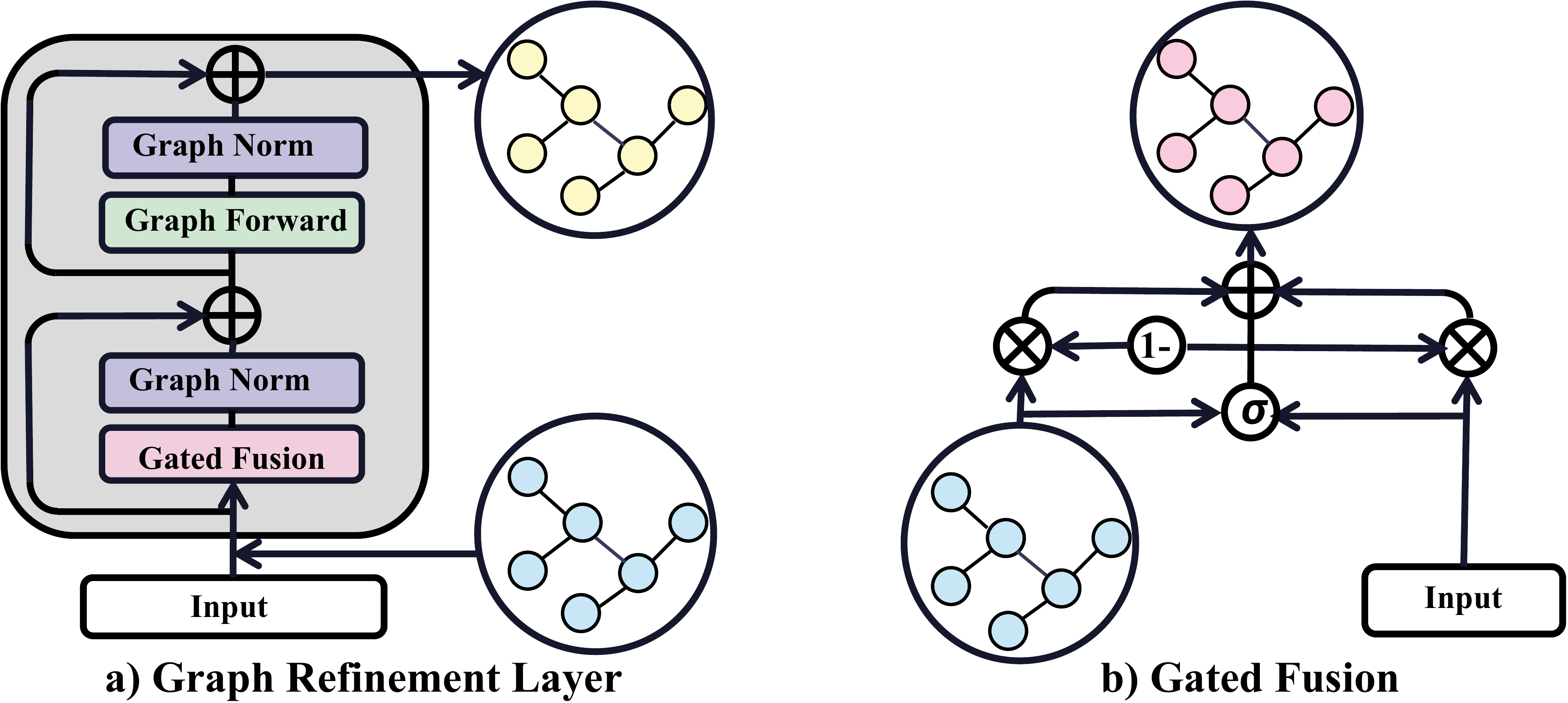}}%{fig/Fig-GRL.pdf}}
%{fig/fig3-v3.png}}
%{fig/graphrefinement.png}}
\vspace{-0.1in}
\caption{The framework Graph Refinement Layer (GRL) module. }
\label{fig3}
\vspace{-0.05in}
\end{figure}

Fig.~\ref{fig3} shows the architecture of the proposed graph refinement layer. Though it is inspired by the encoder of transformer model, there are five major differences between transformer encoder and the newly proposed GRL. a) Transformer encoder captures the temporal features across the sequence, while GRL captures the spatial features and uses the output of transformer encoder to update the local features. b) Transformer encoder only takes hidden vectors as input, while GRL takes both the hidden vectors and graph structures as input. c) Motivated by \cite{dwivedi2020generalization} which demonstrates that batch normalization\cite{ioffe2015batch} is more suitable for graph transformer, GRL replaces layer normalization\cite{vaswani2017attention} with a newly proposed graph normalization, which can be viewed as an extension of batch normalization for graph representation with temporal dependency, to normalize the hidden vectors.
d) GRL replaces multi-head attention in transformer encoder with gated fusion\cite{zheng2020gman} to adaptively fuse the input hidden vectors and the node features in the graph structure. e) GRL further replaces the feed forward in transformer encoder with graph forward to capture the rich spatial features around GPS points.
%which is a stack of $P$ standard GAT modules.

Assume the output hidden vectors at layer $l$ are $\vec{H}_{\tau}^{(l)} \in \mathbb{R}^{l_\tau \times d}$, and the corresponding output graph hidden vectors are $\vec{Z}_{\tau}^{(l)}$, where each sub-graph features $\vec{Z}_{\tau, i}^{(l)} \in \mathbb{R}^{|V_{\tau, i}| \times d}$. Following \cite{vaswani2017attention}, we employ a residual connection for both gated fusion sub-layer and graph forward sub-layer, 
%for each of the two sub-layers 
%\baihua{which two sub-layers? No, gated fusion and graph forward.}, 
followed by graph normalization. That is, the output of each sub-layer is given by $\operatorname{GraphNorm}\left(x + \operatorname{SubLayer}\left(x \right)\right)$, where $\operatorname{SubLayer}$ represents a function that can be either $\operatorname{GatedFusion}$ or $\operatorname{GraphForward}$. $\operatorname{GraphForward}$ is implemented as a stack of $P$ standard GAT modules defined in Eq.~\eqref{eq5} and Eq.~\eqref{eq6}, while $\operatorname{GatedFusion}$ and $\operatorname{GraphNorm}$ are detailed below.

\subsubsection{Gated Fusion} \label{GF}

The hidden-state vectors from the output of transformer encoder layer, i.e. $\vec{Tr}_{\tau}^{(l)}$, capture rich temporal features of the given trajectory, while the graph structure, i.e. $\hat{G}_{\tau}$, captures rich spatial information of each GPS point in the trajectory. Therefore, it is necessary to design a fusion mechanism to adaptively fuse the spatial and temporal features. Inspired by \cite{zheng2020gman}, we propose to use gated fusion to combine the hidden vectors at the $i^{th}$ timestamp, i.e. $tr_{\tau, i}^{(l)} \in \mathbb{R}^{d}$ with the corresponding graph features $\vec{Z}_{\tau, i}^{(l-1)} \in \mathbb{R}^{|V_{\tau, i}| \times d}$:
\begin{eqnarray}
\tilde{Z}_{\tau, i}^{(l)} &=& z_{\tau, i}^{(l)} \odot \hat{tr}_{\tau, i}^{(l)} + \left(1 - z_{\tau, i}^{(l)}\right) \odot \vec{Z}_{\tau, i}^{(l-1)}\nonumber\\
z_{\tau, i}^{(l)} &=& \sigma \left( \hat{tr}_{\tau, i}^{(l)} \mathbf{W}_{z, 1} +  \vec{Z}_{\tau, i}^{(l-1)} \mathbf{W}_{z, 2} + \mathbf{b}_z \right)
\label{eq11} 
%\end{equation}
\end{eqnarray}
Here, $\hat{tr}_{\tau, i}^{(l)} \in \mathbb{R}^{|V_{\tau, i}| \times d}$ repeats $tr_{\tau, i}^{(l)}$ for $|V_{\tau, i}|$ times to ensure that $\hat{tr}_{\tau, i}^{(l)}$ and $\vec{Z}_{\tau, i}^{(l-1)}$ share the same size, $\mathbf{W}_{z, 1}, \mathbf{W}_{z, 2}$ and $\mathbf{b}_z$ represent the learnable weights in the module and $\sigma\left(\cdot \right)$ is the gated activation function, which is implemented as a sigmoid activation function.

\subsubsection{Graph Norm} \label{GN}

Inspired by \cite{dwivedi2020generalization}, we replace layer normalization with graph normalization to normalize graph features within a mini-batch. A mini-batch of input graph structures can be represented as $\{ \hat{G}_{\tau_1}, \hat{G}_{\tau_2}, \cdots, \hat{G}_{\tau_b} \}$ with hidden-state graph features $\{ \tilde{Z}_{\tau_1}^{(l)}, \tilde{Z}_{\tau_2}^{(l)}, \cdots, \tilde{Z}_{\tau_b}^{(l)}\}$, which can be the output of either $\operatorname{GatedFusion}$ sub-layer or $\operatorname{GraphForward}$ sub-layer, where $\tilde{Z}_{\tau}^{(l)}=\{ \tilde{Z}_{\tau, i}^{(l)} \}$ for each $i \in \{1,2,\cdots,l_\tau\}$ and $b$ represents the batch size. 

We first perform mean pooling to obtain graph feature for each sub-graph via Eq.~\eqref{eq12}. 
Here, $M^{(l)}=\{ m_{\tau_i, j}^{(l)} \}$ for each $i \in \{1, 2,\cdots,b\}$ and $j \in \{ 1, 2, \cdots, l_\tau\}$. Note that we assume trajectories $\tau$ in a mini-batch share the same length $l_\tau$. %\baihua{Is this a unrealistic assumption? How to handle it if trajectories have different lengths? Or it is easy to find $b$ trajectories of same length?}
\begin{equation}
m_{\tau_i, j}^{(l)}=\frac{1}{|V_{\tau_i, j}|} \sum\nolimits_{k=1}^{|V_{\tau_i, j}|} \tilde{Z}_{\tau_i, j, k}^{(l)}
\label{eq12}
\end{equation}

We then perform batch normalization on $M^{(l)} \in \mathbb{R}^{b \times l_\tau \times d}$ with the features $\{ \tilde{Z}_{\tau_1}^{(l)}, \tilde{Z}_{\tau_2}^{(l)}, \cdots, \tilde{Z}_{\tau_b}^{(l)}\}$ to get the output $\vec{Z}_{\tau}^{(l)}$:
\begin{equation}
\begin{array}{rl}
    \mu_{\mathcal{B}} = & \frac{1}{b \times l_{\tau}} \sum_{i=1}^{b} \sum_{j=1}^{l_{\tau}} M_{\tau_i, j}^{(l)} \vspace{2ex}  \\
    \sigma_{\mathcal{B}} = & \frac{1}{\sum_{i=1}^{b}\sum_{j=1}^{l_{\tau}} |V_{\tau_i, j}|} \sum_{i=1}^{b}\sum_{j=1}^{l_{\tau}}\sum_{k=1}^{|V_{\tau_i, j}|}  \left( \tilde{Z}_{\tau_i, j, k} - \mu_{\mathcal{B}} \right)^2 \vspace{2ex}  \\
    \widetilde{Z}_{\tau}^{(l)} = & \frac{\tilde{Z}_{\tau}^{(l)} - \mu_{\mathcal{B}}}{\sqrt{\sigma_{\mathcal{B}} + \epsilon}} \vspace{2ex}  \\
    \vec{Z}_{\tau}^{(l)} = & \gamma_{\mathcal{B}} \widetilde{Z}_{\tau}^{(l)} + \beta_{\mathcal{B}}
\end{array}
\label{eq13}
\end{equation}
Here, $\gamma_{\mathcal{B}}$ and $\beta_{\mathcal{B}}$ are learnable parameters for scalar transformation and shift transformation in batch normalization respectively, and $\vec{Z}_{\tau}^{(l)}$ is the output of graph normalization.

\subsection{Transformer Encoder Layer} \label{Trans}

Transformer encoder layer contains a multi-head attention sub-layer and a feed forward sub-layer with a residual connection for each of the two sub-layers, followed by layer normalization. That is, the output of each sub-layer is given by $\operatorname{LayerNorm}\left(x + \operatorname{SubLayer}\left(x \right)\right)$, where $\operatorname{SubLayer}$ represents a function that can be either $\operatorname{MultiHead}$ or $\operatorname{FFN}$.

\subsubsection{Multi-Head Attention}
Given a sequence input $X \in \mathbb{R}^{L \times d}$ with $L$ the length of the sequence, multi-head attention is given by:
%
%\begin{equation}
\begin{eqnarray}
\operatorname{MultiHead\left(Q, K, V\right)} &=& Concat\left(head_1, \cdots, head_h\right)W^O \nonumber \\
%\label{eq14}
%\end{equation}
%
%\begin{equation}
    head_i &=& \operatorname{Attention}\left(QW_i^{Q}, KW_i^{K}, VW_{i}^{V}\right) \nonumber \\
%\label{eq15}
%\end{equation}
%
%\begin{equation}
    \operatorname{Attention\left(Q, K, V\right)} &=& \text{softmax} \left(\frac{QK^T}{\sqrt{d}}\right)V
\label{eq16}
%\end{equation}
\end{eqnarray}
Here, $Q$, $K$ and $V$ refer to the query, the keys and the values for the input $X$ respectively, $W_{i}^{Q}$, $W_{i}^{K}$ and $W_{i}^{V}$ are the learnable parameters of the $i^{th}$ attention head for the query, the keys and the values respectively, $W^O$ is a learnable parameter for the output, and $h$ captures the number of attention heads.

\subsubsection{Feed Forward}

Feed-forward network contains a fully connected network with ReLU activation function:
\begin{equation}
FFN\left(x\right) =\operatorname{ReLU}\left(xW_1 + b_1\right)W_2 + b_2
\label{eq17}
\end{equation}
Here, $W_1$ and $W_2$ are two learnable weights, and $b_1$ and $b_2$ are the bias for the layer.

\subsection{GPSFormer}

GPSFormer first extracts spatial features for each GPS point in the trajectory through the Sub-Graph Generation module introduced in Session~\ref{SG}, and then applies $N$ layers of GPSFormerBlock to capture rich trajectory patterns.

Given a mini-batch of $b$ trajectories $\{ \tau_1, \tau_2, \cdots, \tau_b\}$. We first obtain the initial hidden-state vectors are $\widehat{H}_{\tau}^{(0)}=$ $\{\hat{H}_{\tau_1}^{(0)}$, $\hat{H}_{\tau_2}^{(0)}$, $\cdots$, $\hat{H}_{\tau_b}^{(0)}\}$ $\in \mathbb{R}^{b \times l_\tau \times d}$, accompanied by the initial graph structure $\widehat{G}_{\tau} =$ $\{ \hat{G}_{\tau_1}$, $\hat{G}_{\tau_2}$, $\cdots,\hat{G}_{\tau_b} \}$ and initial graph features $\vec{Z}_{\tau}^{(0)}=$ $\{ \vec{Z}_{\tau_1}^{(0)}$, $\vec{Z}_{\tau_2}^{(0)}$, $\cdots, \vec{Z}_{\tau_b}^{(0)} \}$ though the Sub-Graph Generation module.

After that, we first add position embedding\cite{vaswani2017attention} to $\widehat{H}_{\tau}^{(0)}$ and obtain the input to the transformer encoder:
\begin{equation}
\vec{H}_{\tau}^{(0)} = \widehat{H}_{\tau}^{(0)} + PE\left(\widehat{H}_{\tau}^{(0)}\right)
\label{eq18}
\end{equation}

Next, we input the hidden-state vectors and graph structures to a stacked of $N$ GPSFormer cells:
\begin{equation}
\begin{array}{rl}
\vec{Tr}_{\tau}^{(l)} = & \operatorname{TransformerEncoder}\left( \vec{H}_{\tau}^{(l-1)} \right) \\

\vec{Z}_{\tau}^{(l)} = & \operatorname{GraphRefinement} \left( \vec{Tr}_{\tau}^{(l)}, \vec{Z}_{\tau}^{(l-1)}, \widehat{G}_{\tau} \right) \\

\vec{H}_{\tau}^{(l)} = & \operatorname{GraphReadout} \left(\vec{Z}_{\tau}^{(l)}, \widehat{G}_{\tau} \right)
\end{array}
\label{eq19}
\end{equation}
Here, $\operatorname{TransformerEncoder}$ and $\operatorname{GraphRefinement}$ stand for the transformer encoder layer and the graph refinement layer introduced in Section~\ref{Trans} and Section~\ref{GR} respectively, and $\operatorname{GraphReadout}$ represents graph mean pooling operation similar to Eq.~\eqref{eq12}. The final representation for a mini-batch trajectories is $H_{\tau}^{traj} = \vec{H}_{\tau}^{(N)} \in \mathbb{R}^{b \times l_\tau \times d}$. Besides, we concatenate the mean pooling of $H_{\tau}^{traj}$ with the environmental contexts $f^e_{\tau} \in \mathbb{R}^{b \times f_t}$ (e.g. hour of the day, holiday or not, etc), followed by linear transformation to obtain trajectory-level hidden-state vector $\hat{h}_{\tau}^{traj} \in \mathbb{R}^{b \times d}$.

\subsection{Decoder Model}

Let $H_{\tau}^{traj}$ be the outputs of the encoder model. For simplicity, we denote the representation of a single trajectory as $h^{traj}=\{h_1, h_2, \cdots, h_{l_\tau} \}$. The decoder model uses a GRU model to predict the road segment and moving ratio for each timestamp in the target trajectory. Assume the hidden-state vector in the GRU model at timestamp $t$ is given by $h_{gru}^{(t)}$ with $h_{gru}^{(0)}=\hat{h}_{\tau}^{traj}$. An attention mechanism is adopted to obtain the input of GRU cell, i.e. $a^{(t)}$: 
%\baihua{what does the superscript $j$ here means? $j^{th}$ timestamp? Yes.}
%
\begin{equation}
  a^{(t)} = \sum\nolimits_{i=1}^{l_\tau} \alpha_{i}^{(t)} h_i  \nonumber
\label{eq20}
\end{equation}
\begin{equation}
\begin{array}{rl}
    \alpha_{i}^{(t)} & = \exp \left( \mu_{i}^{(t)} \right) /  \sum_{k=1}^{l_\tau} \exp \left( \mu_{k}^{(t)} \right) \vspace{2ex} \\
    \mu_{i}^{(t)} & = \mathbf{v}^{T} \cdot \operatorname{tanh} \left( \mathbf{W}_g h_{gru}^{(t-1)} + \mathbf{W}_h h_i \right)
\end{array}
\label{eq21}
\end{equation}
Here, $\mathbf{v} \in \mathbb{R}^{d \times 1}$ represents the transformation weight, and $\mathbf{W}_g$ and $\mathbf{W}_h$ are learnable weights in the attention module.

To this end, the hidden-state vectors of the GRU cell are updated by:
\begin{equation}
  h_{gru}^{(j)} = \operatorname{GRU}\left( \left[ x^{(j-1)} \| r^{(j-1)} \| a^{(j)} \right] \right)
\label{eq22}
\end{equation}
Here, $j \in \left[1, 2,\cdots,l_\rho \right]$, $x^{(j-1)}$ and $r^{(j-1)}$ represent the road segment embedding of the predicted road segment and its corresponding moving ratio at the $({j-1})^{th}$ timestamp, 
%$r_{j-1}$ represents the predicted moving ratio at timestamp $t-1$, 
and $\operatorname{GRU}$ represents the GRU cell defined in Eq.~\eqref{eq2}. Finally, $h_{gru}^{(j)}$ is forwarded to a multi-task block to recover the missing trajectory.

\section{Multi-Task Learning for Trajectory Recovery} \label{GCL}

Given a mini-batch of $b$ trajectories, we first use the GPSFormer to obtain the trajectory representation for each sample, i.e., $H_{\tau}^{traj} \in \mathbb{R}^{b \times l_\tau \times d}$, and then forward the hidden-state vectors to the decoder model to obtain the hidden-state vectors of the GRU cell, i.e., $h_{gru}^{(j)}$. For the task of predicting road segment ID, following \cite{ren2021mtrajrec}, we adopt the cross entropy as the loss function:
\begin{equation}
    \mathcal{L}_{id} = -\sum\nolimits_{i=1}^{b} \frac{1}{{l_\rho}} \sum\nolimits_{j=1}^{{l_\rho}}  \log \left(P_{\theta_{enc}, \theta_{dec}}\left( \tilde{e}_{i}^{(j)} | h_{gru}^{(j)}, c_{j} \right)\right) \nonumber
\label{eq23}
\end{equation}
\begin{equation}
    P_{\theta_{enc}, \theta_{dec}}\left( \tilde{e}_{i}^{(j)} | h_{gru}^{(j)}, c_{j} \right) = \frac{\exp \left( h_{gru}^{(j)} \cdot {\mathbf{w}_{i}^{id}} \right) \odot c_{j,\tilde{e}_{i}^{(j)}} }{ \sum_{v \in V} \exp \left( h_{gru}^{(j)} \cdot {\mathbf{w}_{v}^{id}} \right) \odot c_{j,v} } 
\label{eq24}
\end{equation}
Here, $\mathbf{w}^{id} \in \mathbb{R}^{d \times |V|}$ represents a learnable weight and $P_{\theta_{enc}, \theta_{dec}}\left( \tilde{e}_{i}^{(j)} | h_{gru}^{(j)}, c_{j} \right)$ represents the probability of predicting road segment $\tilde{e}_{i}^{(j)}$ at the $j^{th}$ timestamp for the $i^{th}$ trajectory in the mini-batch, given the hidden-state vectors $h_{gru}^{(j)}$ and the constraint mask $c_j$ as defined in the paragraph below, $\tilde{e}_{i}^{(j)}$ represents the ground truth road segment ID at the $j^{th}$ timestamp for the $i^{th}$ trajectory in the mini-batch, and $\theta_{enc}$ and $\theta_{dec}$ are the learnable parameters in the encoder and decoder model respectively.

\paragraph{Constraint Mask Layer} \label{CML}

The goal of the constraint mask layer is to accelerate the convergence of the decoder model and to tackle the difficulties of fine-grained trajectory recovery.
Given a raw GPS trajectory $\tau=\langle \left(p_1, t_1\right), \left(p_2, t_2\right), \cdots, \left(p_{l_\tau}, t_{l_\tau}\right) \rangle$ and the target map-matched $\epsilon_{\rho}$-sample interval trajectory $\rho=\langle \left(q_1, \hat{t}_1\right)$, $\left(q_2, \hat{t}_2\right)$, $\cdots, \left(q_{l_\rho}, \hat{t}_{l_\rho}\right) \rangle$, the constraint mask $c_j \in \mathbb{R}^{l_\rho \times |V|}$ is calculated for each timestamp $\hat{t}_{j}$ in the target trajectory.
%
%we detail the calculation of the constraint mask $c \in \mathbb{R}^{l_\rho \times |V|}$ below. %\baihua{The target is map-matched trajectory or map-matched $\epsilon$-sample interval trajectory?}
%
%For each timestamp $\hat{t}_j$ with $j \in [1, 2, \cdots, l_\rho]$, 
For $\hat{t}_j \in \{ t_1, t_2, \cdots, t_{l_\tau} \}$, the GPS point at timestamp $\hat{t}_j$ is given in the input trajectory, i.e. $t_{k}=\hat{t}_j$ and $q_j=p_k$. Accordingly, we set $c_{j, i} = \omega\left(e_i, p_k\right)$ for each road segment $e_i$ having its distance to $p_k$ within the maximum error of the GPS device (e.g. $100$ meters) and set $c_{j,i}=0$ for other road segments. The function $\omega$ is defined in Eq.~\eqref{eq7}, except that we use another hyper-parameter $\beta$ to replace $\gamma$. For timestamp $\hat{t}_j$ that does not appear in the input trajectory, we set $c_{j,i}=1$ for all road segments $e_i \in V$.

Similarly, we adopt the mean square loss for moving ratio prediction task:
\begin{equation}
    \mathcal{L}_{rate} = \sum\nolimits_{i=1}^{b} \frac{1}{{l_\rho}} \sum\nolimits_{j=1}^{{l_\rho}}  \left( r_{i}^{(j)} - f \left( \left[x_{i}^{(j)} \| h_{gru}^{(j)} \right], \mathbf{w}_{rate} \right) \right)^2
\label{eq25}
\end{equation}
Here, function $f\left( x, y \right) = \sigma\left( x^T \cdot y \right) $, $\mathbf{w}_{rate} \in \mathbb{R}^{2*d \times 1}$ represents a learnable weight and $x_{i}^{(j)}$ represents the road segment embedding for the predicted road segment at the $j^{th}$ timestamp of the $i^{th}$ trajectory in the mini-batch, $r_{i}^{(j)}$ represents the ground truth moving ratio for the $i^{th}$ trajectory in the mini-batch at the $j^{th}$ timestamp, and $\sigma$ refers to the sigmoid function.

To further improve the accuracy of RNTrajRec, we propose a graph classification loss with constraint masks. Given the output graph structure from the last graph refinement layer, i.e. $\vec{Z}_{\tau}^{(N)}$, we calculate the graph classification loss as:
\begin{equation}
    \mathcal{L}_{enc} = -\sum\nolimits_{i=1}^{b} \frac{1}{l_{\tau}} \sum\nolimits_{j=1}^{l_{\tau}}  \log \left(P_{\theta_{enc}}\left( \tilde{e}_{i}^{(j)} | G_{\tau_i, j}, \vec{Z}_{\tau_i, j}^{(N)}\right)\right) \nonumber
\label{eq26}
\end{equation}
\begin{equation}
    P_{\theta_{enc}}\left( \bar{e} | G_{\tau_i, j}, \vec{Z}_{\tau_i, j}^{(N)}\right) = \frac{\exp\left(\vec{Z}_{\tau_i, j, \bar{e}}^{(N)^{T}}  \cdot \mathbf{w} \right) \odot W_{\tau_i, j, \bar{e}} } 
    {\sum_{\bar{v} \in V_{\tau_i, j}} \exp\left(\vec{Z}_{\tau_i, j, \bar{v}}^{(N)^{T}}  \cdot \mathbf{w} \right) \odot W_{\tau_i, j, \bar{v}}}
\label{eq27}
\end{equation}
Here, $\mathbf{w} \in \mathbb{R}^{d \times 1}$ is a learnable weight,  $\vec{Z}_{\tau_i, j, \bar{e}}^{(N)}$ ($W_{\tau_i, j, \bar{e}}$) represents the hidden-state vector (constraint weight) for road segment $\bar{e}$ in the sub-graph $\vec{Z}_{\tau_i, j}^{(N)}$ ($\hat{G}_{\tau_i, j}$), and $\bar{e}_{i}^{(j)}$ represents the ground truth road segment ID at the $j^{th}$ timestamp for the $i^{th}$ input raw GPS trajectory in the mini-batch.

Eq.~(\ref{eq28}) defines the total training loss, where $\lambda_1$ and $\lambda_2$ are hyper-parameters to linearly balance the three loss functions.
\begin{equation}
    \mathcal{L}_{total} = \mathcal{L}_{id} + \lambda_1 \mathcal{L}_{rate} + \lambda_2 \mathcal{L}_{enc}
\label{eq28}
\end{equation}
%

%\begin{table}[]
%\centering
%\caption{Statistics of Datasets}
%\vspace{-0.05in}
%\label{tab:datasets}
% \resizebox{\columnwidth}{!}{%
%\begin{tabular}{c|c|c}
%\hline
%Dataset    &  Number of Trajectories     & Size of Training Area    \\ \hline
%\multirow{2}{*}{ShangHai}   & 2,694,958  & $6.4\operatorname{km} \times 14.4\operatorname{km}$ \\
%                            & Apr 2015   & 9,298 road segments  \\
%\hline
%\multirow{2}{*}{ChengDu}    & 8,302,421 & $8.3\operatorname{km} \times 8.3\operatorname{km}$ \\
%                            & Nov 2016 & 8,781 road segments \\
%\hline
%\multirow{2}{*}{Porto}         & 999,082 & $6.8\operatorname{km} \times 7.2\operatorname{km}$ \\
%                            & Jul 2013-Mar 2014 & 12,613 road segments \\
%\hline
%\end{tabular}
%}
%\vspace{-0.15in}
%\end{table}

\section{Experiment}

\subsection{Experiment Setting}

\subsubsection{Datasets}

Our experiments are based on three real-life trajectory datasets collected from three different cities, namely Shanghai, Chengdu and Porto. Three datasets consist of different number of trajectories collected at various time periods, as listed in Table~\ref{tab:datasets}. 
%Shanghai dataset consists of a total number of 12,694,958 trajectories from Apr. 01 to Apr. 30, 2015. Chengdu dataset consists of a total number of 8,302,421 trajectories from Nov. 01 to Nov. 30, 2016. Porto dataset consists of a total number of 999,082 trajectories from Jul. 01, 2013 to Mar. 20, 2014.  
Since trajectory pattern analysis in urban areas is typically more significant, we keep the central urban area in Chengdu and Porto as the training data, with the size of the selected urban area and the number of road segments covered listed in Table~\ref{tab:datasets}. To demonstrate the scalability of our model, we select a region in Shanghai that includes most suburban areas in addition to the central urban area and hence is much larger than the central urban area, and name this as Shanghai-L dataset. Its size of the selected area and the number of road segments covered  are also listed in Table~\ref{tab:datasets}. Accordingly, we only consider trajectories passing through those selected areas. We want to highlight that considering a central urban area is a common approach used by many existing works~\cite{ren2021mtrajrec,fang2022spatio}. The selected areas, though much smaller than the entire road network, cover most heavy traffics. 
%\baihua{Is it possible to conduct the exp in a larger region for at least one dataset?} \yuqi{I have planned to conduct an exp on Shanghai with 3w road segments containing urban area and suburban area. Large enough? Do we have space to do so?}\baihua{See my messages in Chat.}
%i.e. for Shanghai dataset, we choose an area of size $6.4\operatorname{km} \times 14.4\operatorname{km}$ with $9,298$ road segments; for Chengdu dataset, we choose an area of size $8.3\operatorname{km} \times 8.3\operatorname{km}$ with $8,781$ road segments; and for Porto dataset, we choose an area of size $6.8\operatorname{km} \times 7.2\operatorname{km}$ with $12,613$ road segments. 
We use around $150,000$ trajectories in each dataset for training, and split each dataset into training set, validation set and testing set with a splitting ratio of $7:2:1$. 

\begin{comment}
\begin{figure}
    \centering
    \includegraphics[width=0.45\textwidth]{fig/data.png}
    \caption{Occurrence statistics of each road segment in the Porto dataset with 999,082 trajectories. In the figure, each road segment is color coded based on the total number of trajectories in the training set (in total 999,082 trajectories) that pass the segment, and the blue-colored rectangle bounds the training area selected. Clearly, the training area covers the major traffic (e.g., purple color coded segments refer to the cluster of road segments having the most heavy traffic and most, if not all, purple color coded segments are inside the blue-colored rectangle), which is consistent with the Pareto principle (i.e., 20 percent of the roads are used for 80 percent of the traffic).}
    \label{fig:data}
    \vspace{-0.2in}
\end{figure}
\end{comment}

%The average travel time of a trajectory in Shanghai, Chengdu and Porto datasets is around $697$, $868$ and $783$ seconds respectively. 
To obtain high-sample map-matched trajectories, we use HMM\cite{newson2009hidden} algorithm on original raw GPS trajectories followed by linear interpolation\cite{hoteit2014estimating} to obtain map-matched $\epsilon_\rho$-sample interval trajectory and use the high-sample trajectories with sample interval in the range of $10 \sim 15$ seconds as the ground truth. To obtain low-sample %map-matched
trajectories, we randomly sample trajectory points from high-sample trajectories 
%to obtain low-sample trajectories with 
based on sample interval in the range of $80 \sim 192$ seconds. Specifically, for each dataset, we design a trajectory recovery task which uses only $12.5\%$ or $6.25\%$ points of the given trajectory to recover the remaining $87.5\%$ or $93.75\%$ missing points. Therefore, the average sample interval $\epsilon_\tau$ of the low-sample input trajectories is $8$ or $16$ times higher than that of the origin high-sample trajectories, i.e., $\epsilon_\tau=\epsilon_\rho\times $($8$ or $16$). 
%Besides, to test the robustness of our model, we further design a trajectory recovery task for Chengdu and Shanghai-L dataset that uses only $6.25\%$ points of a given trajectory, i.e., $\epsilon_\tau=\epsilon_\rho\times 16$.

\begin{table}[t]
\centering
\renewcommand{\arraystretch}{1.1} % Default value: 1
\caption{Statistics of Datasets}
\vspace{-0.05in}
\label{tab:datasets}
 \resizebox{\columnwidth}{!}{%
\begin{tabular}{c|c|c|c}
\hline
Dataset & Shanghai-L & Chengdu & Porto \\  
\hline
\# Trajectories & 2,694,958  & 8,302,421 & 999,082 \\
%\# Trajectories for training & 150,058 & 151,256 & 157,516 \\
\# Road segments in training area & 34,986  & 8,781 & 12,613 \\
Size of training area ($\operatorname{km}^2$) & $23.0 \times 30.8$ & $8.3 \times 8.3$ & $6.8 \times 7.2$ \\
Average travel time per trajectory (s) & 699.57 & 868.86 & 783.14 \\
Trajectory collected time & Apr 2015 & Nov 2016 & Jul 2013-Mar 2014 \\
Trajectory raw sample interval (s) & 9.39 & 3.19 & 15.01 \\
Sample interval $\epsilon_\rho$ after processing (s) & 10 & 12 & 15 \\
% Trajectory sample interval (s) & 10 & 12 & 15 \\
\hline
\end{tabular}
}
\vspace{-0.2in}
\end{table}

\subsubsection{Evaluation Metrics}

The task of trajectory recovery is to recover high-sample $\epsilon_\rho$-interval trajectories from low-sample raw trajectories. Following \cite{ren2021mtrajrec}, we adopt both the accuracy of the road segments recovered and the distance error of location inference to evaluate the performances of different models.

\noindent
\textbf{MAE \& RMSE.} 
Mean Absolute Error (MAE) and Root Mean Square Error (RMSE) are common performance metrics typically used for regression tasks. Following \cite{ren2021mtrajrec}, we adopt road network distance to calculate the distance error between two GPS points. That is, given a predicted map-matched trajectory $\hat{\rho}=\langle \left(\hat{e}_1, \hat{r}_1, t_1 \right), \left(\hat{e}_2, \hat{r}_2, t_2 \right),...,\left(\hat{e}_{l_{\rho}}, \hat{r}_{l_{\rho}}, t_{l_{\rho}}  \right) \rangle$, we derive the location of the GPS point $\hat{p}_i$ based on $\hat{e}_i$ and $\hat{r}_i$; similarly, we can find the ground truth GPS point $p_i$ for trajectory $\rho$.
%
%we use \eqref{eq1} to obtain the predicted GPS point, namely $\hat{p}_i$. Similarly, we use \eqref{eq1} to obtain the ground truth GPS point for trajectory $\rho$, namely $p_i$. 
%MAE metric is calculated by 
Accordingly, $MAE\left(\rho, \hat{\rho}\right) = \frac{1}{l_{\rho}} \sum_{i=1}^{{l_{\rho}}} | \operatorname{dist} \left( p_i - \hat{p}_i \right) |$, %while RMSE metric is calculated by 
and $RMSE\left(\rho, \hat{\rho}\right) = \sqrt{\frac{1}{{l_{\rho}}} \sum_{i=1}^{l_{\rho}} \left(\operatorname{dist} \left( p_i - \hat{p}_i \right)\right)^2}$.

\noindent
\textbf{Recall \& Precision \& F1 Score.}
%
%Assume the sequence of $e_i$ in a map-matched GPS trajectory $\rho$ form a travel path in the form of $\{e_1, e_2,...,e_{l_{\rho}}\}$. 
Given a predicted travel path $E_{\hat{\rho}}= \{\hat{e}_1, \hat{e}_2,...,\hat{e}_{l_{\rho}}\}$ extracted from a predicted trajectory $\hat{\rho}$ and the ground truth travel path $E_{\rho}= \{e_1, e_2,...,e_{l_{\rho}}\}$ extracted from the ground truth trajectory $\rho$,
%recall metric is calculated by 
we follow previous work\cite{cui2018personalized, kurashima2010travel, ren2021mtrajrec} and define $Recall\left(\rho, \hat{\rho}\right)=\frac{|E_{\rho} \cap E_{\hat{\rho}}|}{|E_{\rho}|}$, 
%precision metric is calculated by 
and $Precision\left(\rho, \hat{\rho}\right)=\frac{|E_{\rho} \cap E_{\hat{\rho}}|}{|E_{\hat{\rho}}|}$.
%, consistent with previous work\cite{cui2018personalized, kurashima2010travel, ren2021mtrajrec}. 
We also adopt F1 score to further evaluate the models, i.e. $F1\left(\rho, \hat{\rho}\right)=\frac{2 \times Recall\left(\rho, \hat{\rho}\right) \times Precision\left(\rho, \hat{\rho}\right)}{Recall\left(\rho, \hat{\rho}\right) + Precision\left(\rho, \hat{\rho}\right)}$.

\noindent
\textbf{Accuracy.}
The accuracy between predicted trajectory $\hat{\rho}$ and ground truth trajectory $\rho$ is calculated by $Accuracy\left(\rho, \hat{\rho}\right)=\frac{1}{{l_{\rho}}} \sum_{i=1}^{{l_{\rho}}} \mathbf{1}\{ e_i=\hat{e}_i \}$, which evaluates the model's ability to match the recovered GPS trajectory to the corresponding road segments.

\noindent
\textbf{SR\%k.}
In order to further evaluate the robustness of the models, we create another task, namely elevated road recovery, to evaluate models' ability to accurately recover trajectories on elevated roads and nearby trunk roads. As mentioned in \cite{iland2018rethinking}, GPS locations can be inaccurate in densely populated and highly built-up urban areas, making the task of trajectory recovery more challenging and significant. 
As we foresee the recovered trajectories in those areas contain more errors, we propose to use $SR\%k$ instead of average F1 score that is more sensitive to poor cases. Specifically, given a predicted travel path $E_{\hat{\rho}}$ and ground truth travel path $E_{\rho}$, we choose a sub-trajectory $\widehat{E}_{\rho}$ of length $\hat{l}_{\rho}$ which drives on or near an elevated road along with the predicted sub-trajectory $\widehat{E}_{\hat{\rho}}$. Then, $SR\%k$ calculates the proportion of trajectories with $F1(\widehat{E}_{\rho}, \widehat{E}_{\hat{\rho}})$ value exceeding $k$ to evaluate the models' ability to discriminate complex roads using contextual trajectory information.

% Since the road network structure can be extremely complicated around the urban elevated road, we propose to use $SR\%k$ instead of average F1 score which is more sensitive to bad cases for evaluation. 

\subsubsection{Parameter Settings}

In our experiments, we implement RNTrajRec and all the baseline models in Python Pytorch\cite{paszke2019pytorch} framework. We set the size of hidden-state vectors $d$ to $512$ for Chengdu and Porto datasets. Due to memory limitation, we set $d$ to $256$ for Shanghai-L dataset. We use the same size of hidden-state vectors for all the models across datasets. Also, we set the size of grid cell to $50m \times 50m$. For our model, we set both the number of GAT modules in road network representation $M$ and the number of GPSFormer layers $N$ to $2$, and set the number of GAT modules in graph refinement layer $P$ to $1$. Also, we set hyper-parameters $\delta$ and $\gamma$ to $400$ and $30$ meters respectively, set the number of attention heads $h$ in both Eq. \eqref{eq6} and Eq. \eqref{eq16} to $8$ and $\lambda_1, \lambda_2$ in Eq. \eqref{eq28} to $10$ and $0.1$ respectively. The size of $f_r$ is set to $11$, i.e. $8$ for level of road segment, $1$ for length of road segment, $1$ for number of in-edges and $1$ for number of out-edges, and that for $f_l$ is set to $25$, i.e. $24$ for one-hot vector for hour of the day and $1$ for holiday or not. Following \cite{ren2021mtrajrec}, we set hyper-parameter $\beta$ (used to set constraint mask) to $15$ meters. All the models are trained with Adam optimizer\cite{kingma2014adam} for $30$ epochs with batch size $64$ and learning rate $10^{-3}$. All the experiments are conducted on a machine with AMD Ryzen 9 5950X 16 cores CPU and 24GB NVIDIA GeForce RTX 3090 GPU.

\subsubsection{Compared models}

\begin{table*}[htbp]
\renewcommand\tabcolsep{3.3pt} % 调整表格列间的宽度
\renewcommand{\arraystretch}{1.2} % Default value: 1
\caption{Performance evaluation for different methods on trajectory recovery task.}
\vspace{-0.15in}
\begin{center}
\begin{tabular}{|c|c|cccccc|cccccc|}
\hline
\multicolumn{2}{|c|}{\multirow{2}*{Method}}&\multicolumn{6}{c|}{Chengdu ($\epsilon_\tau=\epsilon_\rho*8$)}&\multicolumn{6}{c|}{Chengdu ($\epsilon_\tau=\epsilon_\rho*16$)} \\
\multicolumn{2}{|c|}{~} & Recall & Precision & F1 Score & Accuracy & MAE & RMSE & Recall & Precision & F1 Score & Accuracy & MAE & RMSE  \\
\hline
\multicolumn{2}{|c|}{Linear + HMM} & 0.6597 & 0.6166 & 0.6351 & 0.4916 & 358.24 & 594.32 & 0.4821 & 0.4379 & 0.4564 & 0.2858 & 525.96 & 760.47 \\
\multicolumn{2}{|c|}{DHTR + HMM} & 0.6385 & 0.7149 & 0.6714 & 0.5501 & 252.31 & 435.17 & 0.5080 & 0.6930 & 0.5821 & 0.4130 & 325.14 & 511.62 \\
\hline
\multirow{7}{*}{\hspace{0.05cm}\rotatebox{90}{End-to-End Methods}\hspace{0.05cm}} & t2vec + Decoder & 0.7123 & 0.7870 & 0.7441 & 0.5601 & 194.29 & 307.22 & 0.6490 & 0.7725 & 0.7013 & 0.4627 & 254.29 & 375.16 \\
& Transformer + Decoder & 0.7365 & 0.8229 & 0.7742 & 0.5902 & 177.13 & 287.33 & 0.6091 & 0.7187 & 0.6537 & 0.4258 & 294.73 & 420.91 \\
& MTrajRec & 0.7565 & 0.8410 & 0.7938 & 0.6081 & 160.29 & 261.11 & 0.6643 & 0.7957 & 0.7202 & 0.4918 & 231.84 & 348.55 \\
& T3S + Decoder & 0.7535 & 0.8394 & 0.7913 & 0.6092 & 163.58 & 263.42 & 0.6634 & 0.7838 & 0.7144 & 0.4897 & 234.00 & 352.35 \\
& GTS + Decoder & 0.7514 & \uline{0.8428} & 0.7917 & 0.6105 & 157.83 & \uline{254.51} & 0.6569 & 0.7900 & 0.7131 & 0.4825 & 231.78 & 344.21  \\
& NeuTraj + Decoder & \uline{0.7608} & 0.8405 & \uline{0.7961} & \uline{0.6152} & \uline{156.25} & 254.70 & \uline{0.6644} & \uline{0.7979} & \uline{0.7213} & \uline{0.4942} & \uline{227.19} & \uline{341.03}  \\
%\cline{2-14}
%& RNTrajRec w/o GRL & 0.7696 & 0.8773 & 0.8177 & 0.6459 & 144.61 & 240.22 & 0.6776 & \textbf{0.8585} & 0.7540 & 0.5260 & 204.28 & 312.47 \\
%& RNTrajRec w/o GCL & 0.7773 & 0.8744 & 0.8209 & 0.6472 & 140.59 & 236.49 & 0.6871 & 0.8523 & 0.7578 & 0.5364 & 197.13 & 307.03  \\
& RNTrajRec (Ours) & \textbf{0.7831} & \textbf{0.8812} & \textbf{0.8272} & \textbf{0.6609} & \textbf{132.69} & \textbf{219.20} & \textbf{0.6926} & \textbf{0.8573} & \textbf{0.7632} & \textbf{0.5413} & \textbf{195.91} & \textbf{304.52} \\
\hline
\end{tabular}

\vspace{0.5em}
\hspace{0em}
\begin{tabular}{|c|c|cccccc|cccccc|}
\hline
\multicolumn{2}{|c|}{\multirow{2}*{Method}}&\multicolumn{6}{c|}{Porto ($\epsilon_\tau=\epsilon_\rho*8$)}&\multicolumn{6}{c|}{Shanghai-L ($\epsilon_\tau=\epsilon_\rho*16$)} \\
\multicolumn{2}{|c|}{~} & Recall & Precision & F1 Score & Accuracy & MAE & RMSE & Recall & Precision & F1 Score & Accuracy & MAE & RMSE  \\
\hline
\multicolumn{2}{|c|}{Linear + HMM} & 0.5837 & 0.5473 & 0.5629 & 0.3624 & 175.00 & 284.16 & 0.6055 & 0.5633 & 0.5801 & 0.3825 & 383.25 & 555.68  \\
\multicolumn{2}{|c|}{DHTR + HMM} & 0.5578 & 0.6837 & 0.6118 & 0.4250 & \uline{104.41} & \uline{168.83} & 0.5144 & 0.6533 & 0.5696 & 0.3974 & 308.72 & 454.87  \\
\hline
\multirow{7}{*}{\hspace{0.05cm}\rotatebox{90}{End-to-End Methods}\hspace{0.05cm}} & t2vec + Decoder & 0.6543 & 0.7546 & 0.6977 & 0.4738 & 124.77 & 184.57 & 0.6397 & 0.7487 & 0.6831 & 0.4544 & 298.35 & 420.22  \\
& Transformer + Decoder  & 0.6343 & 0.7449 & 0.6816 & 0.4590 & 132.70 & 195.10 & 0.5850 & 0.7039 & 0.6306 & 0.4160 & 357.46 & 496.25 \\
& MTrajRec & 0.6449 & 0.7504 & 0.6905 & 0.4656 & 125.81 & 184.94 & 0.6106 & 0.7372 & 0.6603 & 0.4328 & 327.32 & 456.40 \\
& T3S + Decoder & 0.6392 & 0.7377 & 0.6816 & 0.4551 & 131.12 & 191.99 & 0.6282 & 0.7408 & 0.6721 & 0.4510 & 303.55 & 428.35 \\
& GTS + Decoder & 0.6474 & \uline{0.7612} & 0.6967 & 0.4761 & 118.07 & 173.77 & \uline{0.6441} & \uline{0.7809} & \uline{0.6987} & \uline{0.4714} & \uline{276.23} & \uline{391.74} \\
& NeuTraj + Decoder & \uline{0.6544} & 0.7558 & \uline{0.6984} & \uline{0.4808} & 119.45 & 176.27 & 0.6337 & 0.7472 & 0.6787 & 0.4542 & 293.65 & 414.59 \\
%\cline{2-14}
%& RNTrajRec w/o GRL & 0.7618 & \textbf{0.8757} & 0.8111 & 0.6602 & 112.99 & 192.99 & 0.6671 & 0.7946 & 0.7227 & 0.5145 & 101.51 & 150.17 \\
%& RNTrajRec w/o GCL & 0.7737 & 0.8744 & 0.8175 & 0.6654 & 117.27 & 190.79 & 0.6683 & 0.7928 & 0.7227 & 0.5119 & 102.13 & 152.10 \\
& RNTrajRec (Ours) & \textbf{0.6778} & \textbf{0.7950} & \textbf{0.7293} & \textbf{0.5230} & \textbf{97.66} & \textbf{145.87} & \textbf{0.6663} & \textbf{0.8294} & \textbf{0.7332} & \textbf{0.5145} & \textbf{229.74} & \textbf{335.19} \\
\hline
\end{tabular}
\label{tab1}
\vspace{-0.7cm}
\end{center}
\end{table*}

To evaluate the effectiveness of RNTrajRec,
%GPSFormer, 
we implement in total eight baselines. 
%
%\begin{itemize}
%    \item 
i) \textbf{Linear} \cite{hoteit2014estimating}\textbf{+HMM} %\cite{song2012quick} 
uses linear interpolation to obtain a high-sample trajectory, and then HMM algorithm to obtain a map-matched $\epsilon_\rho$-sample interval trajectory.
%    \item 
ii) \textbf{DHTR+HMM} %\cite{song2012quick} 
%\cite{wang2019deep} 
replaces Linear with a hybrid Seq2Seq model with kalman filter\cite{kalman1960new}.
%to recover missing GPS points in the trajectory, and then HMM algorithm to obtain map-matched $\epsilon_\rho$-sample interval trajectory.
%    \item 
iii) \textbf{t2vec} 
%\cite{li2018deep} 
proposes a deep learning network for trajectory similarity learning with a  BiLSTM\cite{hochreiter1997long} model to capture the temporal dependency of the given trajectory.
 %   \item 
iv) \textbf{Transformer}\cite{vaswani2017attention} learns the representation with temporal dependency. 
%is a novel framework for representation learning with temporal dependency.
Follow \cite{ren2021mtrajrec}, we use grid cell index and time index for each sample point in the trajectory. 
%We use the same input as \cite{ren2021mtrajrec} which contains a grid cell index and time index for each sample point in the trajectory.
%    \item
v) \textbf{NeuTraj} %\cite{yao2019computing} 
adds a spatial attention memory model to LSTM %model, namely SAM, 
to capture rich nearby spatial features for trajectory representation.
%    \item
vi) \textbf{T3S} 
%\cite{yang2021t3s} %proposes a learning-based method that 
combines a self-attention network with a spatial LSTM model to capture spatial and structural features of trajectories.
%    \item 
vii) \textbf{GTS}\cite{han2021graph} is the state-of-the-art method for learning trajectory similarity in spatial network which uses POIs as input.  
%It proposes a novel graph-based framework. However, it uses POI instead of GPS points as input. 
To adapt GTS to our problem setting, we regard each intersection in the road network as a POI, %. Meanwhile, for road segments with length longer than $100$ meters,
and cut every long road segment into segments every $100$ meters  and treat each cut point as a POI. Also, we use the embedding vector of the nearest POI to obtain the representation of each GPS point in the trajectory. 
%    \item 
viii) \textbf{MTrajRec} 
%\cite{ren2021mtrajrec} 
is the state-of-the-art method for trajectory recovery task.
%which consists of a GRU\cite{cho2014properties} module for encoding trajectory and a map-constraint decoder with multi-task learning for trajectory recovery task. 
%\end{itemize}

\textit{Remark 1}: Traj2SimVec is a novel model for trajectory similarity learning with auxiliary supervision for sub-trajectory. However, its encoder model 
%for Traj2SimVec 
is simply a RNN-based model, which is similar to t2vec.
%\cite{li2018deep}. 
Since we mainly compare the performance of different models on GPS trajectory representation learning, we do not include this method for comparison. 

\textit{Remark 2}: We refer to \textbf{A+Decoder} as using the encoder model proposed in \textbf{A} and the decoder model proposed in \cite{ren2021mtrajrec} for trajectory recovery task.

\subsection{Performance Evaluation}

We report the experimental results in Table \ref{tab1}. Numbers in bold font indicate the best performers, and numbers underlined represent the best performers among existing baselines without considering our model.
%\baihua{Yuqi, can we remove underline and change double underline to underline? It's relatively easy to find the better performances when we compare the two variants of our model.}
As shown, RNTrajRec outperforms all baseline models on all the datasets. 

In particular, Linear+HMM performs the worst on all the datasets and its performance drops significantly as the sample interval becomes longer, e.g. the accuracy drops from $0.4916$ to $0.2858$ when the sample interval increases from $96$ seconds to $192$ seconds on Chengdu dataset. We also observe that DHTR+HMM outperforms Linear+HMM, which confirms that linear interpolation is not suitable for recovering low sample GPS trajectories. 
%linear interpolation method used in Linear + HMM. 
%%%What's more, as \cite{feng2018deepmove, ren2021mtrajrec} mentioned, it is more reliable to model ID sequence than numerical sequence. Therefore, it is not surprising that t2vec+Decoder performs the worst on Chengdu and Shanghai datasets among almost all learning-based models that take raw GPS sequence as input to the RNN (i.e. BiLSTM) model. Overall,

Another observation is that NeuTraj and GTS, two models proposed for GPS trajectory learning, outperform MTrajRec when they include a decoder model proposed in \cite{ren2021mtrajrec} on top of their encoder models. This well demonstrates that these two models are able to capture spatio-temporal information in low-sample trajectories. 
%Besides, T3S+Decoder outperforms MTrajRec on Chengdu and Shanghai datasets.
%
%two out of three models for GPS trajectory learning (i.e. NeuTraj and GTS) outperform MTrajRec when a decoder model proposed in \cite{ren2021mtrajrec} is added on top of these encoder models, which shows that these models are able to capture spatio-temporal information in low-sample trajectories.

It is observed that end-to-end methods perform better than two-stage methods. 
Among existing end-to-end models, we observe that NeuTraj achieves the highest accuracy on Chengdu and Porto datasets, while GTS significantly outperforms NeuTraj on Shanghai-L dataset. We believe that it is due to the complex road network structure in Shanghai-L that leads to the high performance for graph-based method. In addition, our method consistently outperforms all these models with a large margin. 
To be more specific, as compared with the best baseline, RNTrajRec improves F1 score and accuracy by an average of $4.85\%$ and $8.48\%$ respectively, and reduces MAE and RMSE by an average of $27.42$ and $35.91$ meters respectively for Chengdu dataset; it improves F1 score and accuracy by $4.94\%$ and $9.14\%$ respectively and reduces MAE and RMSE by $46.49$ and $56.55$ meters respectively for Shanghai-L dataset; it improves F1 score and accuracy of the best baseline by $4.42\%$ and $8.78\%$ respectively and reduces MAE and RMSE by $6.75$ and $22.96$ meters respectively on Porto dataset. 
%
%For Chengdu dataset, RNTrajRec improves F1 score and accuracy of the best baseline by an average of $3.65\%$ and $4.64\%$ respectively, and reduce MAE and RMSE by an average of $27.42$ and $35.91$ meters respectively. For Shanghai dataset, RNTrajRec improves F1 score and accuracy of the best baseline by $4.52\%$ and $5.02\%$ respectively; and reduce MAE and RMSE by $22.39$ and $31.82$ meters respectively. For Porto dataset, our newly proposed RNTrajRec can improve F1 score and accuracy of the best baseline by $3.09\%$ and $4.22\%$ respectively. MAE and RMSE are reduced by $6.75$ and $22.96$ meter respectively. 
This clearly proves the effectiveness of RNTrajRec for trajectory recovery, which is mainly contributed by following two reasons. 
%The reasons for the improvement lie in two aspects. 
Firstly, RNTrajRec pays attention to the important road network information and road network structure around each GPS point in the trajectory. Secondly, several novel modules are proposed in this paper, such as GridGNN and GPSFormer, that enable the model to learn rich spatial and temporal features of the given trajectory.

\subsection{Additional Experiment Results}

We conducts additional experiments on two new datasets, namely Shanghai and Chengdu-Few, to evaluate the performance of RNTrajRec in various data distributions.

For Shanghai dataset, we select another area in Shanghai with a total number of $9,298$ road segments, covering an area of $6.4 \times 14.4 \operatorname{km}^2$, and meanwhile keep the other settings (e.g., trajectory sample interval time) of original Shanghai-L dataset. Experiment result shown in Table \ref{tab:extend} verifies that RNTrajRec still performs the best among all the baseline models.

Because of its design, RNTrajRec might benefit more from a training dataset that contains a large number of trajectories, as compared to other baselines. In order to demonstrate that our model is equally competitive even when the training set contains much fewer trajectories, we construct a new dataset, namely Chengdu-Few, that randomly samples only 20,000 trajectories (roughly 20\% of the origin Chengdu dataset) from the original training dataset in Chengdu and meanwhile keep the other settings (e.g., \# of road segments and size of the training area) of original Chengdu dataset. In such a setting, our model needs to learn the representation of each road segment along with the representation of each subgraph with limited trajectories. 

As reported in Table~\ref{tab:extend}, our method outperforms all the baseline in Chengdu-Few dataset and we believe the reasons lie in the following two aspects. Firstly, when the training set has a limited number of trajectories, many road segments appear only a few times in the training dataset. However, when graph neural network updates a certain road segment, the features of the surrounding road segments are also updated, which offers opportunities for each road segment to learn a good feature representation with limited number of trajectories. Secondly, the input features of each subgraph are obtained from the representation of the road network directly. Therefore, as long as each road segment in the road network has a well-trained feature representation, it's possible for each subgraph to obtain a good feature representation.

Last but not the least, we find that although our model still achieves the best results on the new Chengdu-Few dataset, the improvement is slightly less significant than that achieved under the original Chengdu dataset, e.g., on Chengdu dataset, our model outperforms the best baseline by 7.42\% in accuracy and 5.80\% in F1 score; on Chengdu-Few dataset, these two numbers drop to 6.57\% in accuracy and 2.75\% in F1 score. We think the main reason is that a transformer model typically requires a large amount of training data. Consequently, when the amount of training data is insufficient, the quality of the final representation will be affected (to a certain degree).

\begin{table*}[htbp]
\renewcommand\tabcolsep{3.3pt} % 调整表格列间的宽度
\renewcommand{\arraystretch}{1.2} % Default value: 1
\vspace{0.05in}
\caption{Performance evaluation on additional Shanghai and Chengdu-Few datasets.}
\begin{center}
\begin{tabular}{|c|c|cccccc|cccccc|}
\hline
\multicolumn{2}{|c|}{\multirow{2}*{Method}}&\multicolumn{6}{c|}{Shanghai ($\epsilon_\tau=\epsilon_\rho*8$)}&\multicolumn{6}{c|}{Chengdu-Few ($\epsilon_\tau=\epsilon_\rho*8$)} \\
\multicolumn{2}{|c|}{~} & Recall & Precision & F1 Score & Accuracy & MAE & RMSE & Recall & Precision & F1 Score & Accuracy & MAE & RMSE  \\
\hline
\multicolumn{2}{|c|}{Linear + HMM} & 0.7563 & 0.7158 & 0.7329 & 0.5730 & 205.82 & 331.43 & 0.6597 & 0.6166 & 0.6351 & 0.4916 & 358.24 & 594.32 \\
\multicolumn{2}{|c|}{DHTR + HMM} & 0.6682 & 0.7728 & 0.7123 & 0.5876 & 160.31 & 261.17 & 0.5729 & 0.6938 & 0.6243 & 0.4940 & 282.52 & 468.91  \\
\hline
\multirow{7}{*}{\hspace{0.05cm}\rotatebox{90}{End-to-End Methods}\hspace{0.05cm}} & t2vec + Decoder & 0.6914 & 0.7155 & 0.6965 & 0.5295 & 184.42 & 280.36 & 0.6591 & 0.7690 & 0.7055 & 0.5069 & 237.70 & 364.77 \\
& Transformer + Decoder & 0.7249 & 0.7702 & 0.7404 & 0.5786 & 161.03 & 253.22 & 0.6542 & 0.7569 & 0.6977 & 0.5051 & 237.07 & 364.29  \\
& MTrajRec & 0.7417 & 0.7874 & 0.7581 & 0.5924 & 148.26 & 232.41 & \uline{0.7026} & \uline{0.8083} & \uline{0.7483} & \uline{0.5418} & \uline{206.08} & \uline{320.35}  \\
& T3S + Decoder & 0.7581 & 0.7923 & 0.7695 & 0.6009 & 145.58 & 231.79  & 0.6984 & 0.7964 & 0.7405 & 0.5374 & 207.97 & 322.31  \\
& GTS + Decoder & 0.7525 & \uline{0.8144} & \uline{0.7766} & \uline{0.6172} & \uline{134.58} & \uline{212.78} & 0.6888 & 0.8074 & 0.7396 & 0.5312 & 207.48 & 321.80  \\
& NeuTraj + Decoder & \uline{0.7588} & 0.7976 & 0.7726 & 0.6058 & 141.63 & 223.26 & 0.6973 & 0.7912 & 0.7378 & 0.5403 & 211.64 & 330.74  \\
& RNTrajRec (Ours) & \textbf{0.7824} & \textbf{0.8735} & \textbf{0.8218} & \textbf{0.6674} & \textbf{112.19} & \textbf{180.96} & \textbf{0.7205} & \textbf{0.8309} & \textbf{0.7689} & \textbf{0.5774} & \textbf{179.74} & \textbf{287.84} \\
\hline
\end{tabular}
\label{tab:extend}
\end{center}
\end{table*}

\subsection{Robustness Study}

\begin{comment}
\begin{table}[htbp]
\vspace{-0.6cm}
\renewcommand\tabcolsep{3.3pt} % 调整表格列间的宽度
\renewcommand{\arraystretch}{1.2} % Default value: 1
\caption{Performance evaluation for different methods on elevated road trajectory task.}
\vspace{-0.07in}
\centering
\begin{tabular}{|c|ccccc|}
\hline
\multirow{2}{*}{Method}&\multicolumn{5}{|c|}{Chengdu ($\epsilon_\tau=\epsilon_\rho*8$)} \\
& SR\%90 & SR\%80 & SR\%70 & SR\%60 & SR\%50 \\
\hline
Linear + HMM & 0.0673 & 0.1694 & 0.2520 & 0.3170 & 0.3946 \\
DHTR + HMM & 0.0942 & 0.2873 & 0.4878 & 0.6132 & 0.6863 \\
\hline
t2vec + Decoder & 0.1232 & 0.4051 & 0.6791 & 0.8133 & 0.8922 \\
Transformer + Decoder & 0.1492 & 0.4536 & 0.7141 & 0.8465 & 0.9061  \\
MTrajRec & 0.1554 & 0.4495 & 0.7018 & 0.8453 & 0.9196  \\
T3S + Decoder & 0.1691 & 0.4995 & 0.7491 & 0.8650 & 0.9234 \\
GTS + Decoder & 0.1748 & 0.4883 & 0.7445 & 0.8675 & 0.9249 \\
NeuTraj + Decoder & \uline{0.1882} & \uline{0.5140} & \uline{0.7619} & \uline{0.8783} & \uline{0.9307} \\
\hline
RNTrajRec w/o GRL & 0.2066 & 0.5432 & 0.7873 & 0.8916 & 0.9421 \\
RNTrajRec w/o GCL & 0.2033 & 0.5489 & 0.7890 & 0.8974 & 0.9425  \\
RNTrajRec & \textbf{0.2534} & \textbf{0.5880} & \textbf{0.8097} & \textbf{0.9094} & \textbf{0.9505} \\
\hline
\end{tabular}
\vspace{-0.3cm}
\label{tab2}
\end{table}
\end{comment}

\begin{figure}
    \centering
    \vspace{-0.2cm}
    \includegraphics[width=0.48\textwidth]{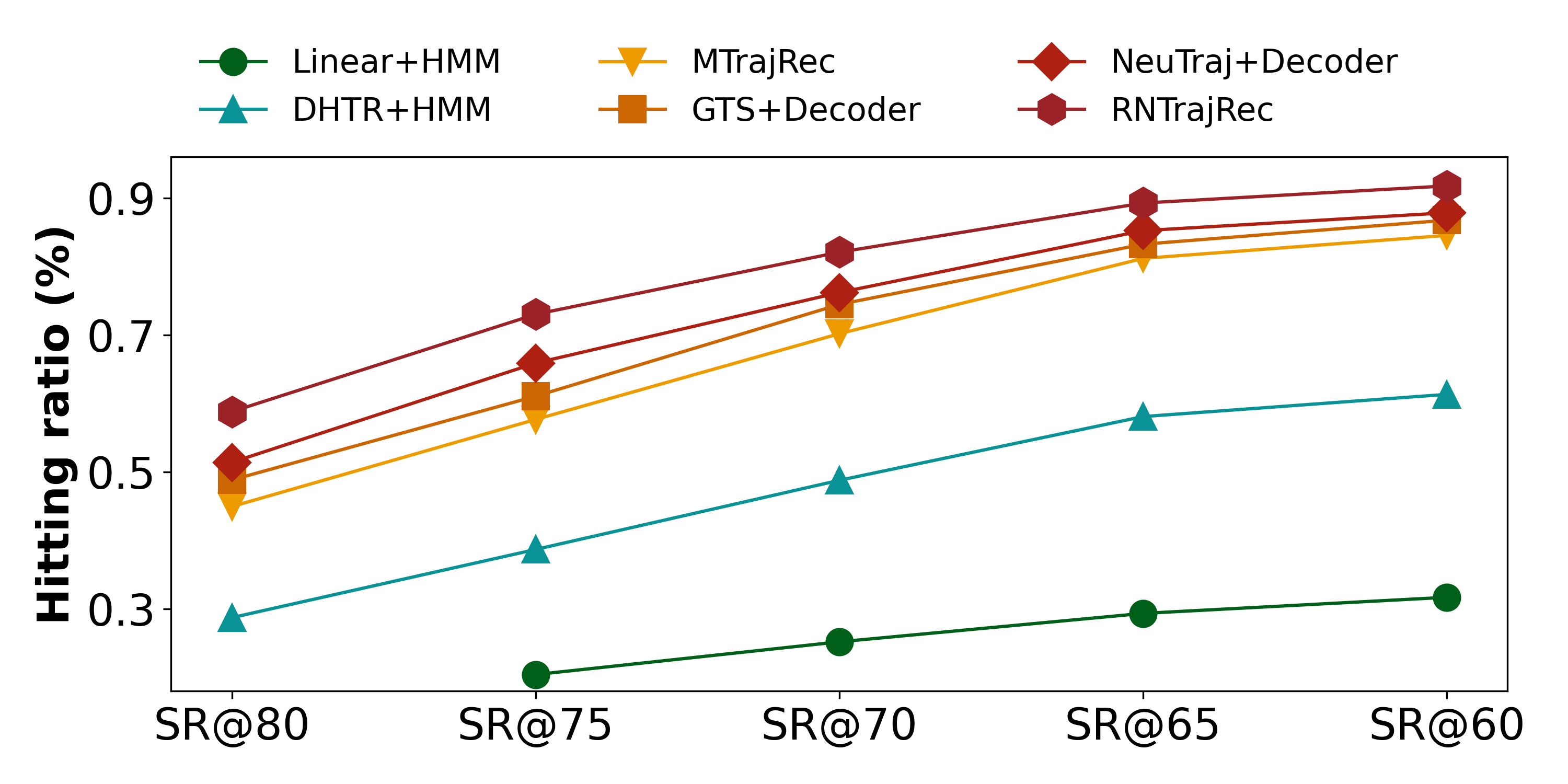}
    \vspace{-0.3cm}
    \caption{Performance evaluation for different methods on elevated road trajectory task (Chengdu: $\epsilon_\tau=\epsilon_\rho*8$).}
    \label{fig:robust}
    \vspace{-0.5cm}
\end{figure}

Fig.~\ref{fig:robust} reports the experimental result for elevated road trajectory task on Chengdu dataset. We design this task to serve two main purposes. 
%The aim of designing this task lies in two aspects. 
First, elevated roads and nearby trunk roads normally experience high traffic volume and their traffic conditions are typically more complicated (e.g., elevated roads are more likely to experience traffic congestion during peak hours). 
%
%people often travel through elevated roads and nearby trunk roads, and the traffic conditions on these roads are more complicated. For example, during morning rush hour and evening rush hour, some elevated roads will become congest. 
Therefore, evaluating the models' ability to discover these sophisticated spatio-temporal patterns is significant. Second, the road network structure around the elevated road is more complex than other urban roads. For example, there are usually two-way trunk road segments under the elevated road segments, which brings greater challenges to the recovery of the road segments on these elevated roads. 

%As listed in Table ~\ref{fig:robust}, 
As shown in Fig.~\ref{fig:robust}, RNTrajRec significantly outperforms all the baseline models. Interestingly, all the learning-based models significantly outperform HMM-based methods (i.e. Linear/DHTR+HMM), which shows the inability of HMM to discover these complex patterns. In addition, our model can achieve a high F1 score ($>0.8$) for $58.8\%$ trajectories, which outperforms the best baseline by $14.4\%$. %$7.4\%$. 
%\baihua{What are the numbers inside the brackets? 0.2534-0.1882=0.0652} 
%What's more, RNTrajRec outperforms all its variants by a large margin. Therefore, we conclude that our model is more robust than existing baseline models.

\begin{figure*}
    \centering
    \vspace{-0.15in}
    \includegraphics[width=0.95\textwidth]%{fig/Fig-Casev4.pdf}
    {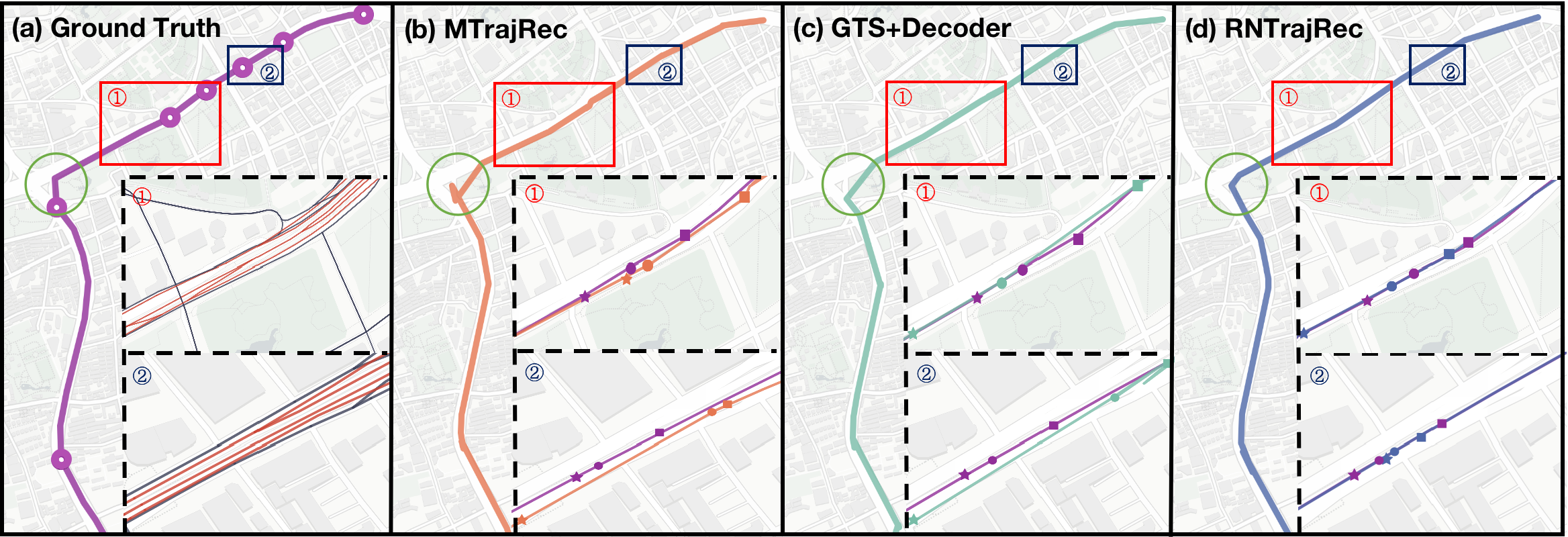}
    \caption{A case study for trajectory recovery. Purple circles in (a) represent the sample points in an input low sample trajectory. The red/black lines in dash-line rectangles of (a) represent the partial elevated/main road segments. The figures in the lower right corner of (b), (c), and (d) plot the detailed recovery results corresponding to the areas inside the red/black rectangles and the
    marker shapes indicate the sample timestamps (e.g., a purple star and an orange star in the dash-line rectangle labelled \textcircled{2} in (b) represent a point recovered by MTrajRec and the corresponding ground truth point of the same timestamp.)
    %markers of the same shape (star, circle, and diamond) indicate the recovered sample point and the ground truth point of the same timestamp.}
    %\baihua{Two sample points are inside the red rectangle as shown in Fig4(a). Three points are recovered in Fig4(b) - (c). Based on the sampling interval (e.g., $\epsilon_\tau=\epsilon_\rho*8$), shall we have more points? In addition, red rectangle in (a) is different from those in (b)-(d) in terms of position.}\yuqi{I just sample some typical recovered point in the target trajectory, not necessarily continuous and not necessarily all the points in the area. For the position, I have fixed it.}
    }
    \label{fig:case}
    \vspace{-0.1in}
\end{figure*}

\subsection{Efficiency Study}

\begin{figure}[htbp]
    \centering
    \includegraphics[width=0.48\textwidth]{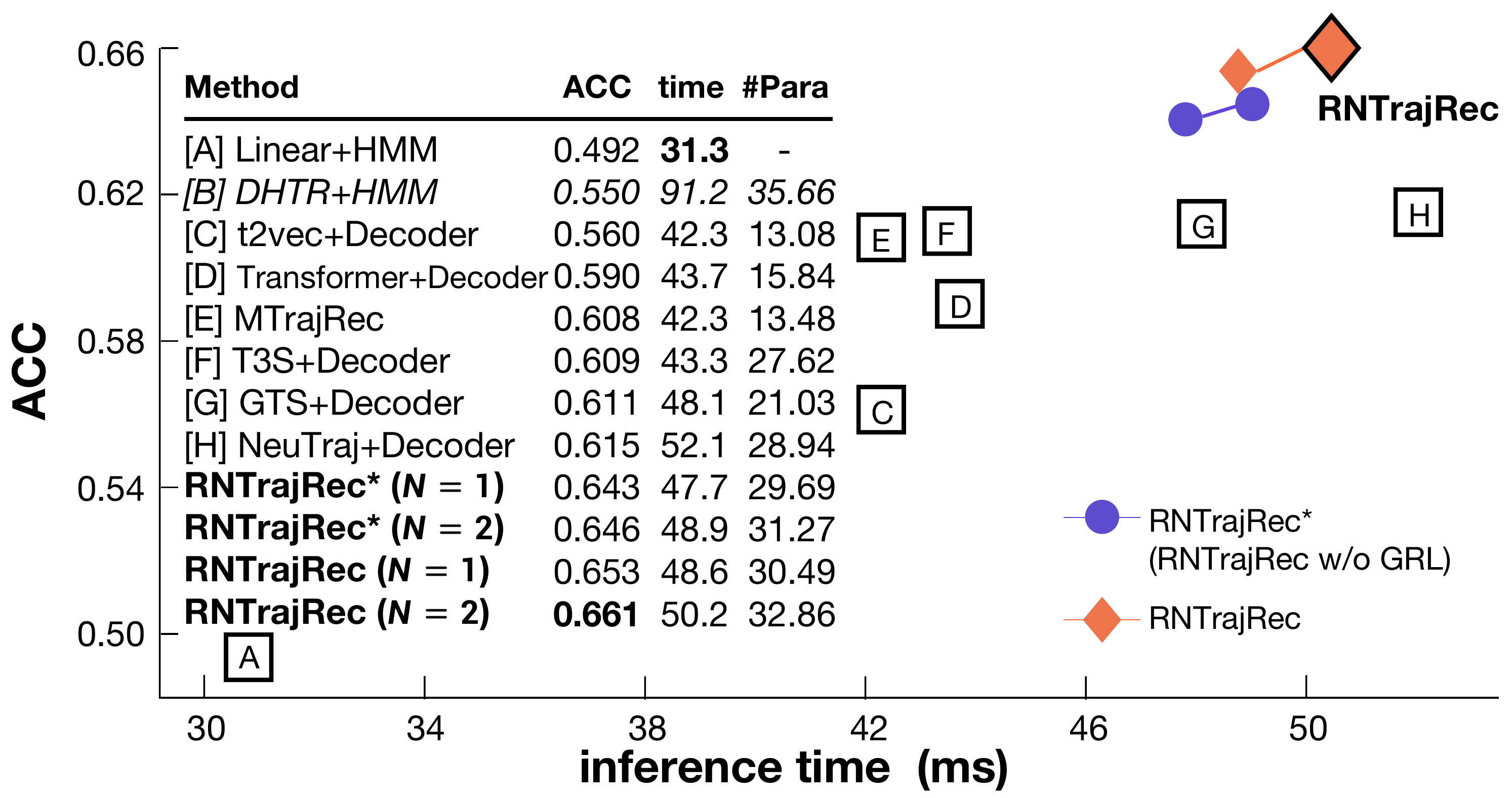}
    \vspace{-0.35cm}
    \caption{Efficiency study %for different models 
    on Chengdu dataset ($\epsilon_\tau=\epsilon_\rho*8$)}.
    \vspace{-0.7cm}
    \label{fig:effi}
\end{figure}

In addition to the effectiveness of different methods, we also evaluate their efficiency by considering two different aspects, including inference time that refers to the time required to recover a trajectory during the inference phase and the number of parameters. 
%Specifically, inference time refers to the time required to recover a trajectory during the inference phase, which is an important factor that should be taken into consideration in real-life applications.
%
As shown in Fig.~\ref{fig:effi}, RNTrajRec requires less time during the inference than NeuTraj+Decoder, and comparable time cost against GTS+Decoder when using only one layer of GPSFormer. However, our model significantly outperforms both NeuTraj+Decoder and GTS+Decoder in terms of accuracy even if we set $N=1$. Besides, DHTR+HMM is observed to spend the most time during inference, probably due to the inefficiency of the adopted kalman filter module. Another observation is that Linear+HMM requires the least time during inference, however it suffers from low accuracy for trajectories with low sample rate. In short, RNTrajRec requires $50.2$ microseconds to recover a low-sample trajectory, which is efficient and practical in practice. 

\textit{Remark:} The time spent to obtain the road network representation via GridGNN is excluded from the inference time reported above. This is because the road network representation can be learned in advance as it is independent of the input trajectory used for the inference task. 
%should not be taken into account, since its result is independent of the input trajectory, it is feasible to save the feature representation of each road segment in advance.

\subsection{Case Study}

Fig.~\ref{fig:case} visualizes trajectories recovered by different models, 
%RNTrajRec and two baseline models MTrajRec and GTS+Decoder respectively, 
where the input trajectory is a low-sample elevated road
trajectory. % from our datasets. 
We sample partial underlying road network structures (e.g., elevated road segments represented by red lines and main road segments represented by black lines) in the two dash-line rectangles of Fig.~\ref{fig:case}(a). As we can observe from the visualization, the road network near the elevated roads is complicated. 
%with the same low-sampling-rate trajectory data at a keep ratio $12.5\%$. 
Purple lines in figures represent the ground truth trajectory and orange, green, blue-colored trajectories represent the trajectory recovered by MTrajRec, GTS+Decoder, and RNTrajRec respectively. %From the figure, it can be found
We can observe that the trajectory recovered by our model matches the ground truth much better (e.g., the trajectories in the green circles provide one example), as RNTrajRec is able to capture the spatial and temporal patterns of trajectories in a more accurate manner. 
%the GPS position recovered of both 
%MTrajRec and GTS+Decoder both fail to recover the missing points between the two input GPS points shown in the green-circled area of Fig.~\ref{fig:case}(a), while RNTrajRec is able to recover the missing points more accurately. \yuqi{What do you mean by mentioning "the missing points between the two input GPS points"? Actually, I would like to say that the trajectory shapes in the green circle are different, and our trajectory is more closer to the ground truth, therefore, our model can capture more accurate spatial and temporal pattern (maybe speed pattern, like the moving speed is typically slower when meeting intersection).}
%is not accurate enough at the turning place, while the recovered GPS location is accurate for RNTrajRec. 
We further plot snapshots of two sections of restored trajectories (bounded by rectangles labelled \textcircled{1} and \textcircled{2}) by different models in the elevated road using the small pictures in the lower right corner in Fig.~\ref{fig:case} (b)-(d), together with three recovered points from each section and their corresponding ground truth as examples. It can be observed that 
%there is an error between
both road sections restored by MTrajRec and those restored by GTS+Decoder deviate from the ground truth,
%\baihua{Can we circle the errors in Fig.4 (b) and (c).}\yuqi{How to circle?}\baihua{I mean circle the parts of orange/green lines that are different from the purple lines, e.g., the part near the intersection?}\yuqi{It is really hard to circle since there are too much error for the orange line. I have upload the PPT file in figure-ICDE-cyq.pptx.}, 
while the road sections restored by our model match the ground truth trajectory perfectly.
In addition, we want to highlight that points recovered by the two baseline models lack spatial consistency due to their insufficient use of road network. For example, the orange/green star in Fig.~\ref{fig:case} (b/c)-\textcircled{2} is a point on the main road, while the next recovered point (i.e., the orange/green circle) is located on the elevated road. Although the two points seem to be located on the same road in our visualization, 
%which cannot be observe from our visualization. 
the shortest path distance between those two points is larger than $2000$ metres, implying that the recovered path between these two points is very different from the ground truth. 
%, significantly larger than their Euclidean distance that we can observe from our visualization.
%In addition, we would like to highlight that given a GPS point $p_1$ on a main road and another GPS point $p_2$ on a nearby elevated road, their shortest path distance along the road network is typically large, significantly larger than their Euclidean distance that we can observe from our visualization. For example, a recovered point that is 100 metres away from its real location at a main road actually introduces an error of close to 3000 metres to its shortest distance to another point on a nearby elevated road. 
In other words, 
%
%shortest path distance between a GPS point on an elevated road segment and the main road under the elevated road is typically large
%\baihua{cannot really get the point. What do you mean?}\yuqi{See the message.}, 
%due to the connectivity of the road network, therefore
the fact that our model can recover more accurate trajectories for elevated roads with complex network topology is significant.

\begin{table*}[htbp]
\renewcommand\tabcolsep{3.3pt} % 调整表格列间的宽度
\renewcommand{\arraystretch}{1.2} % Default value: 1
\caption{Ablation studies on Chengdu and Porto datasets.}
\vspace{-0.1in}
\centering
\begin{tabular}{|c|cccccc|cccccc|}
\hline
\multirow{2}{*}{Variants}&\multicolumn{6}{c|}{Chengdu ($\epsilon_\tau=\epsilon_\rho*8$)} & \multicolumn{6}{c|}{Porto   ($\epsilon_\tau=\epsilon_\rho*8$)} \\
%Variants & Recall & Precision & F1 Score & Accuracy & MAE & RMSE \\
& Recall & Precision & F1 Score & Accuracy & MAE & RMSE & Recall & Precision & F1 Score & Accuracy & MAE & RMSE \\
\hline
w/o GRL & 0.7696 & 0.8773 & 0.8177 & 0.6459 & 144.61 & 240.22 & 0.6671 & 0.7946 & 0.7227 & 0.5145 & 101.51 & 150.17 \\
w/o GF & 0.7725 & 0.8765 & 0.8191 & 0.6439 & 141.31 & 234.28 & 0.6697 & 0.7926 & 0.7234 & 0.5133 & 102.07 & 151.04 \\
w/o GAT & 0.7821 & 0.8729 & 0.8229 & 0.6292 & 144.70 & 237.82 & 0.6747 & \textbf{0.7962} & 0.7279 & 0.5195 & 98.65 & 147.98 \\
w/o GN & 0.7827 & 0.8672 & 0.8200 & 0.6306 & 146.56 & 241.25 & 0.6729 & 0.7951 & 0.7264 & 0.5171 & 99.84 & 148.05 \\
w/o GCL & 0.7773 & 0.8744 & 0.8209 & 0.6472 & 140.59 & 236.49 & 0.6683 & 0.7928 & 0.7227 & 0.5119 & 102.13 & 152.10 \\
\hline
RNTrajRec & \textbf{0.7831} & \textbf{0.8812} & \textbf{0.8272} & \textbf{0.6609} & \textbf{132.69} & \textbf{219.20} & \textbf{0.6778} & 0.7950 & \textbf{0.7293} & \textbf{0.5230} & \textbf{97.66} & \textbf{145.87} \\
\hline
\end{tabular}
\vspace{-0.15in}
\label{tab:ablation}
\end{table*}

\subsection{Ablation Study}

To further prove the effectiveness of the modules 
proposed in the paper, we create five variants of RNTrajRec. \textbf{w/o GRL} replaces the graph refinement layer (GRL) in GPSFormer with standard transformer layer and ignores the graph structure input, i.e. $\widehat{G}_{\tau}$ and $\vec{Z}_{\tau}^{(0)}$; \textbf{w/o GF} replaces gated fusion discussed in Session~\ref{GF} with concatenation operation and feed forward network; \textbf{w/o GN} replaces graph normalization discussed in Session~\ref{GN} with standard layer normalization in transformer encoder; \textbf{w/o GAT} employs feed forward network but not graph attention network in graph refinement layer; \textbf{w/o GCL} removes graph classification (GCL) loss defined in Eq.~\eqref{eq26}.  Experimental results are listed in Table \ref{tab:ablation}. We can observe that RNTrajRec consistently outperforms all its variants, which proves the significance of these modules.

As mentioned in Section~\ref{SG}, the local surrounding graph structure of GPS points is important for understanding the movement of a trajectory. Therefore, we observe a significant drop in overall performance after removing GRL,
%the graph refinement layer, 
especially for Recall and F1 Score. Besides, we observe that RNTrajRec w/o GRL significantly outperforms Transformer+Decoder, with the input to the transformer layer being their only difference. Therefore, we conclude that a well-designed input embedding is significant for complex encoding models like transformer. 

GRL is considered the most important component in RNTrajRec, in which graph normalization, gated fusion and graph forward is replaced with layer normalization, multi-head attention and feed forward network in original transformer respectively. In order to better justify the design of these individual modules, we design ablation study to answer the following three questions.

$\bullet$ Q1: Why we use graph normalization to replace layer normalization in graph refinement layer?

$\bullet$ Q2: Why we use graph attention network to replace feed forward network in graph refinement layer?

$\bullet$ Q3: Why we use gate fusion instead of other simple method (say concatenation or MLP) in graph refinement layer?

\textbf{Q1: Why Graph Normalization?} The idea of designing the graph normalization is inspired by \cite{dwivedi2020generalization}, in which an experiment has been conducted to demonstrate that batch normalization is more suitable for graph transformers. Therefore, we design a similar normalization strategy for dynamic graph transformer (or spatial-temporal transformer). 
%Besides, we believe that the reason behind the design of graph normalization is that 
In addition, we observe that layer normalization ignores other nodes in the same sub-graph, which may cause the normalized features of different nodes in the same sub-graph lack discrimination. 

\textbf{Q2: Why Graph Attention Network?} There are two main reasons behind the design of an additional graph attention network in graph refinement layer. First, as mentioned in Session IV-C, each subgraph $\vec{Z}_{\tau, i}^{(0)}$ directly uses the features in the road segments as the input but totally ignores the relationship inside each subgraph.
%s is not taken into consideration. 
Second, transformer encoder discussed in Session IV-E can effectively aggregate features from the entire trajectory, therefore, the relationship in each subgraph can be refined with the output of transformer encoder.

\textbf{Q3: Why Gated Fusion?} As mentioned in Session IV-D, we adopt  gated fusion to adaptively fuse the input hidden
vectors and the node features in the graph structure. 

Extensive results reported in Table~\ref{tab:ablation} well verify that the performance of RNTrajRec will degrade if any of the three parts are replaced/removed.
%Graph normalization and gated fusion are the two key components proposed in GRL. Again, we observe a significant drop in overall performance after removing these components. 
%graph refinement layer. 
%We observe that graph normalization is particularly significant, because 
%We believe that the reason behind this is that 
%layer normalization ignores other nodes in the same sub-graph and hence may cause the normalized features of different nodes in the same sub-graph lack discrimination. 

As for GCL, it is mainly to guide the process of trajectory encoding so as to generate more accurate trajectory representation. As shown in Table \ref{tab:ablation}, we observe that GCL indeed improves the accuracy of the recovered trajectory.

In addition, we study the effectiveness of different road network representation methods. As shown in Fig.~\ref{fig4}(a), we compare GridGNN with three novel graph neural networks (GNNs), namely GCN, GIN and GAT
%\cite{kipf2016semi}, GIN\cite{xu2018powerful} and GAT\cite{velivckovic2017graph} 
for graph representation. All these models are implemented using standard DGL\cite{wang2019deep2} library. We observe that GridGNN consistently performs the best, which shows the effectiveness of integrating grid information for road network representation. In addition, 
%among the three existing GNNs, 
GAT outperforms GIN and GCN for graph representation, which shows the significance of self-attention mechanism.

\subsection{Parameter Analysis}

\subsubsection{The impact of the number of GPSFormerBlock $N$}

The number of GPSFormerBlock $N$ directly influences the complexity of RNTrajRec. To examine its impact, we vary $N$ in RNTrajRec from $1$ to $5$ and report the results in Fig.~\ref{fig4}(b). 
Note, when $N$ is too large, the model will be more prone to overfitting, resulting in a decrease in its performance. We observe from the results that RNTrajRec achieves the highest accuracy when $N=3$.  However, considering the efficiency of the model and the usage of GPU memory, we set $N$ to $2$.
%in our previous experiments.

\subsubsection{The influence of receptive field of GPS point $\delta$}

The value of receptive field of GPS point $\delta$ decides the size of the surrounding sub-graph of each GPS point in the trajectory.
%when we form the sub-graph for each GPS point
A larger $\delta$ allows the sub-graph to capture more information of surrounding area with a higher memory cost. In order to study the impact of $\delta$ on RNTrajRec's performance, we vary $\delta$ from $100$ meters to $800$ meters. 
%With the increase of $\delta$, each GPS point can gather more information about the road network structure, however the memory used for storing these graph structures and graph features increase dramatically. 
As shown in Fig.~\ref{fig4}(c), RNTrajRec achieves the highest accuracy when $\delta$ is set to $600$ meters. 
%However, RNTrajRec achieves a comparable performance when $\delta=400$ meters.  
%
%there is no significant improvement as $\delta$ increases from $400$ meter to $600$ meter. In order 
To balance between the effectiveness and efficiency of the model, we set $\delta$ to $400$ meters.

\subsubsection{The influence of hyper-parameter $\gamma$ in Eq. \eqref{eq7}}

\begin{figure}[t]
\vspace{-0.1cm}
\centerline{\includegraphics[width=9cm]{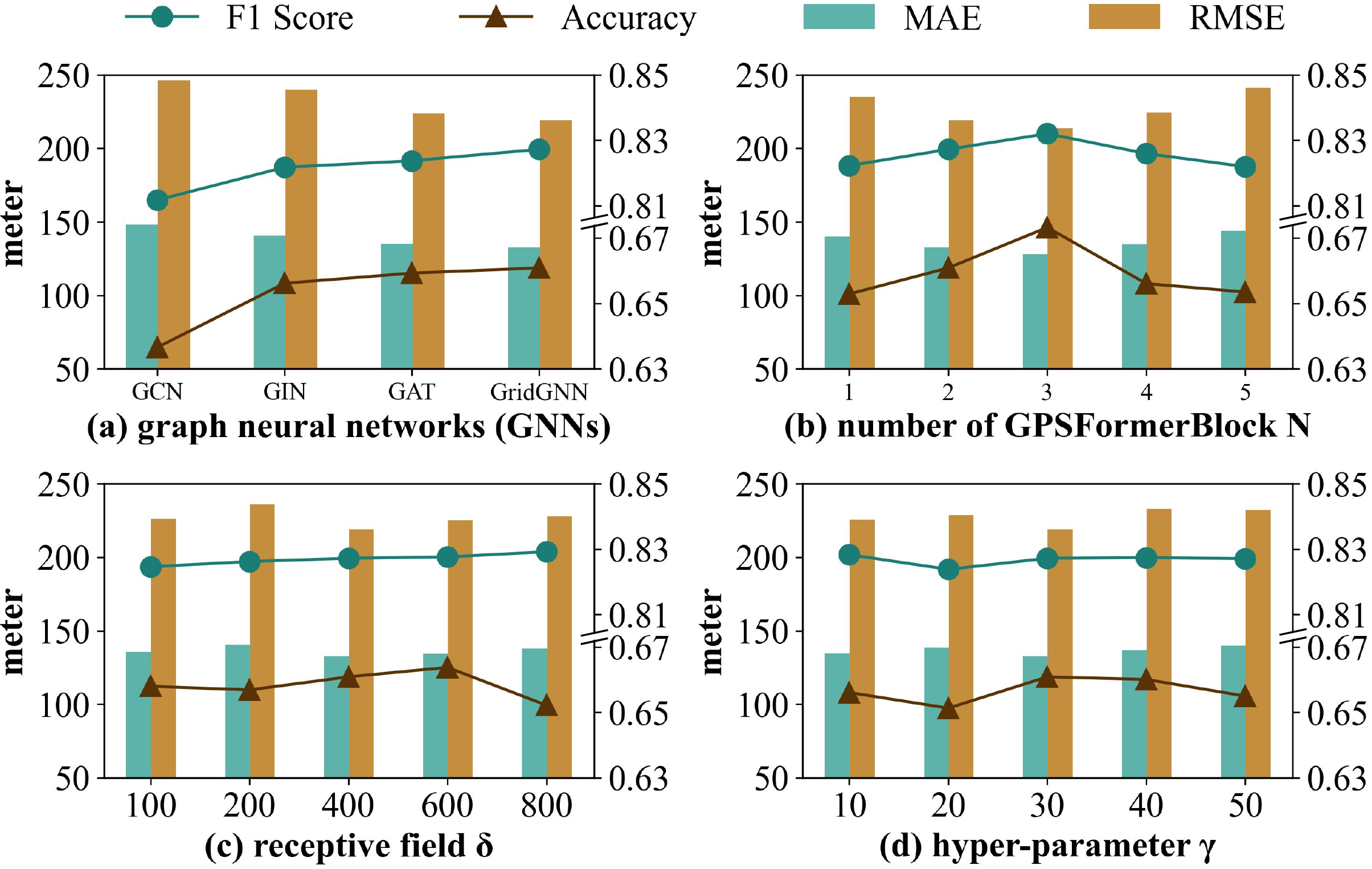}}
\vspace{-0.15in}
\caption{The performance of RNTrajRec on Chengdu dataset ($\epsilon_\tau=\epsilon_\rho*8$) under different hyper-parameters.}
\label{fig4}
\vspace{-0.6cm}
\end{figure}

We vary the hyper-parameter $\gamma$ from $10$ meters to $50$ meters. With the increase of $\gamma$, the initial hidden-state vectors $H_{\tau}^{(0)}$ will pay more attention to the road segments that are closer to the GPS point, while pay less attention to those far away from the GPS point. Because of the uncertainty of GPS errors, the effect of increasing $\gamma$ on the model is also uncertain. 
%When $\gamma$ is very large, the model can solve simple problems more easily (i.e. the case when GPS point is close to the target road segment). However, the model will get confused when the GPS point is far away from the target road segment. When $\gamma$ is small, the model will pay more attention to the surrounding road segments. However, it will also influence the convergence of the model. 
As shown in Fig.~\ref{fig4}(d), we observe that the performance of the model does not vary much as $\gamma$ changes. We believe the main reason is that GPSFormer can dynamically adjust the weight of each node in the sub-graph for the current GPS point, making the model insensitive to the changes of hyper-parameter $\gamma$.

\subsection{Discussion}

We have explored many different techniques during the design of RNTrajRec. Though most of our attempts are effective, there is one technique that we tried but did not work. 
%One thing that we tried but did not work is that when designing the graph refinement layer, %we not only want 
When designing the graph refinement layer, ideally the graph transformer network is able to refine not only the graph embedding of each subgraph but also the weight of each node in every subgraph. Specifically, we try to use the refined graph embedding at layer $l$, i.e. $\vec{Z}_{\tau}^{(l)}$, to obtain a new weight for each node by first transforming $\vec{Z}_{\tau}^{(l)}$, followed by either sigmoid function or softmax function. However, we find out that both sigmoid function and softmax function perform worse than a simple approach that does not refine the weight in each subgraph, i.e. directly using mean pooling as discussed in Section~\ref{GR}. We believe the main reason behind this failed attempt is that the linear transformation is too simple %that it is difficult 
to learn valuable weights without proper supervision. 
%Again we will include this as ``lesson learned'' in the extended version. 

\section{Conclusion}

Trajectory recovery is a significant task for utilizing low-sample trajectories effectively. In this paper, we propose a novel spatial-temporal transformer-based model, namely RNTrajRec, to capture rich spatial and temporal information of the given low-sample trajectory. 
%Specifically, we propose a GridGNN module for road network representation and a GPSFormer module for encoding GPS trajectory.
Specifically, we propose a 
%grid-partitioned 
road network representation module, namely GridGNN and a novel spatial-temporal transformer module, namely GPSFormer for encoding a GPS trajectory.
%We obtain the spatial topology of each GPS point through a Sub-Graph Generation module.
We then forward the hidden-state vectors to a multi-task decoder model
%to perform road segment classification and moving ratio regression 
to recover the missing GPS points. 
%To further improve the accuracy, 
Also, we propose a graph classification loss with constraint mask to guide the process of trajectory encoding. 
%GridRNN incorporate a LSTM model for modeling grid sequence and a graph neural network to capture graph topology. 
%Given a trajectory, we first use Sub-Graph Generation module to obtain the sub-graph structure around each GPS point. Then, the proposed GPSFormer module takes in both the hidden-state vectors and sub-graph structures for trajectory encoding. Finally, we forward the hidden-state vectors into a multi-task decoder model to perform road segment classification and moving ratio regression and recover the missing GPS points. To further improve the accuracy, we propose a graph classification loss with constraint mask to guide the process of trajectory encoding. 
%To the best of our knowledge, we are the first to combine road network representation with GPS trajectory representation. 
Extensive experiments on three real-life datasets show the effectiveness and efficiency of the proposed method.

\section{Acknowledgements}

This research is supported in part by the National Natural Science Foundation of China under grant 62172107.

\normalem
%\IEEEtriggeratref{55}
\bibliographystyle{IEEEtran}
\bibliography{reference}

\end{document}